\documentclass{article} 
\usepackage{iclr2027_conference,times}

\usepackage{amsmath,amsfonts,bm}

\def\eqref#1{equation~\ref{#1}}

\def\1{\bm{1}}

\DeclareMathAlphabet{\mathsfit}{\encodingdefault}{\sfdefault}{m}{sl}
\SetMathAlphabet{\mathsfit}{bold}{\encodingdefault}{\sfdefault}{bx}{n}

\usepackage{todonotes}
\usepackage{hyperref}
\usepackage{url}
\usepackage{bbm}
\usepackage{graphicx}
\usepackage{subcaption}
\usepackage{wrapfig}
\usepackage{booktabs}
\usepackage{amsmath,amssymb}
\usepackage{array}
\usepackage{makecell}
\usepackage[capitalize,noabbrev]{cleveref}

\newlength{\betasweepheight}
\newlength{\casadisweepheight}
\newlength{\trainingcurveheight}

\newtheorem{theorem}{Theorem}[section]

\newtheorem{assumption}[theorem]{Assumption}

\title{Sufficiency of Zeroth-Order Reward Shaping for Policy Gradient in Stabilization Control}

\author{
Yisheng Zhang$^{1,2}$ \quad
Tao Wang$^{1}$ \quad
Sicun Gao$^{1}$\thanks{
Corresponding author.
Email: \texttt{sicung@ucsd.edu}.
}
\\
$^{1}$University of California San Diego, La Jolla, CA, USA
\\
$^{2}$Tsinghua University, Beijing, China
\\
\texttt{zhang-ys23@mails.tsinghua.edu.cn, taw003@ucsd.edu}
}

\iclrfinalcopy
\begin{document}

\maketitle
\lhead{Sufficiency of Zeroth-Order Reward Shaping for Policy Gradient in Stabilization Control}

\begin{abstract}

Reward shaping is fundamental to modern robotic control with deep reinforcement learning (RL), yet practitioners still rely heavily on heuristic principles borrowed from classical optimal control and trajectory optimization.
Existing methods rarely distinguish reward terms that are intrinsic to the control objective from numerical regularizers, leading to brittle hyperparameter tuning.
To determine which quantities a reward must contain, we study the stabilization control problem with a focus on zeroth-order (configuration) and first-order (velocity) information.
We theoretically and empirically demonstrate that policy gradient methods can successfully solve stabilization tasks without first-order reward terms, adding such terms can instead introduce severe sensitivity as their scale grows. Conversely, our findings confirm that reward functions must be zeroth-order complete over goal-relevant coordinates, while the first-order state remains necessary in the policy observation under our low-dissipation assumptions.
Overall, these results provide actionable and principled guidance for reward design in robotic RL.
\end{abstract}

\section{Introduction}

Reward shaping lies at the core of deep reinforcement learning (RL) methods, particularly in modern robotic control. Despite its central role, current practice lacks a systematic study. Researchers and practitioners predominantly rely on heuristic principles inherited from traditional frameworks such as optimal control~\citep{bellman1957dynamic,pontryagin1962mathematical,kalman1960contributions,bryson1975applied} and trajectory optimization~\citep{jacobson1970differential,hargraves1987direct,betts1998survey,posa2014direct,kelly2017trajectory}.
For instance, locomotion and physics-based motion policies are commonly trained with weighted sums of pose, velocity, torque, action-rate, and smoothness terms \citep{peng2018deepmimic,hwangbo2019learning,lee2020learning,rudin2022learning}.
Crucially, existing methodologies fail to distinguish whether a specific reward component is intrinsically required by the underlying control problem or merely serves as an optimization regularizer imposed by the solver. This ambiguity hinders principled reward design and complicates hyperparameter tuning in deep RL.

Unlike classical approaches, RL offers a distinct route to control: learning a policy from experience rather than solving a known system model \citep{kober2013reinforcement, sutton2018reinforcement}. As a model-free method, an RL agent receives sampled trajectories and scalar returns rather than derivatives of the physical transition or reward map.
Consequently, reward terms that are physically intuitive can produce poorly scaled value targets or rare, extreme returns---factors to which deep RL is notoriously sensitive \citep{mnih2015human,vanhasselt2016learning,henderson2018deep,engstrom2020implementation,andrychowicz2021what}.
Conversely, this distinction suggests that regularity-related reward terms essential for gradient-based classical methods may be unnecessary for RL. These observations naturally lead to our central question:
$$\textit{What quantities are strictly required to appear in the reward function?}$$
To answer this, we study stabilization control, a broad class of problems in robotics, and focus specifically on velocity terms within the reward function. Standard reward shaping typically applies quadratic penalties to both zeroth- and first-order states (i.e., configurations and velocities).
Although velocity terms can regularize Newton-based trajectory optimization, they often introduce brittleness into RL. Our analyses show that policy gradient methods solve varied stabilization problems without first-order reward terms. We further find that such terms increase hyperparameter sensitivity as their scale grows, making training more prone to failure and helping explain why reward clipping is effective in practice.

Moreover, we prove that velocity terms may be unnecessary in the reward yet remain essential in the observation: removing velocities from the state space significantly degrades performance. We also demonstrate that reward functions must be zeroth-order complete over all goal-relevant configuration coordinates.
Overall, this work aims to provide systematic understanding and principled guidance for reward shaping in deep RL for robotic control.

Our contributions are summarized as follows:
\begin{itemize}
  \item We demonstrate that velocity reward terms in stabilization control problems are required by classical gradient-based approaches (e.g., trajectory optimization), but are not intrinsically required by policy gradient RL.

  \item We conduct extensive experiments showing that omitting velocity reward terms preserves performance and improves robustness, while velocity observations remain necessary under low-dissipation dynamics.

  \item We provide insights into the practical interaction between zeroth- and first-order reward terms, offering a principled explanation for the efficacy of common reward shaping heuristics like reward clipping.
\end{itemize}

\section{Related Work}

\paragraph{Velocities in classical control and trajectory optimization.}

The modern optimal-control problem emerged from dynamic programming and the maximum principle, which posed control as the optimization of cumulative cost subject to dynamics \citep{bellman1957dynamic,pontryagin1962mathematical}. Kalman's linear-quadratic theory then connected a quadratic cost to a closed-form linear feedback law \citep{kalman1960contributions,bryson1975applied,lewis2012optimal}.
For a mechanical state $s=[q,\dot q]^\top$ and an action $a$, the state determines what the controller can distinguish, whereas blocks of $Q$ and $M$ encode which deviations the designer prefers to suppress using $s^\top Q s+a^\top Ma$. Penalizing velocity consequently became a convenient way to favor damping, regular motion, or tracking around a nominal trajectory, leveraging the velocity information already embedded in the system dynamics.

Nonlinear trajectory optimization preserved this separation while changing how trajectories are computed. Shooting and differential dynamic programming propagate dynamics and use local quadratic models \citep{jacobson1970differential,todorov2005generalized,tassa2012synthesis}; direct transcription and collocation optimize discretized states and controls under equality constraints \citep{hargraves1987direct,betts1998survey,kelly2017trajectory}; and contact-implicit formulations incorporate impacts and friction into the optimization \citep{posa2014direct}.
While software such as CasADi has made automatic differentiation and sparse nonlinear programming formulations broadly accessible \citep{andersson2019casadi}, a positive-definite state-cost matrix $Q$ is often used to add curvature to the objective in trajectory optimization.

\paragraph{Reward shaping and trajectory-based learning in deep RL.}

Reward shaping can accelerate exploration, but unrestricted shaping can also change the optimal policy. Potential-based transformations are a notable invariant class \citep{ng1999policy}, recently unified with intrinsic motivation under a Bayes-adaptive MDP formulation \citep{lidayan2025bamdp}. Thus, adding more physically interpretable terms need not make a reward easier to learn from.
Modern robotic RL scaled this recipe with deep policies and massively parallel simulation \citep{schulman2017proximal,makoviychuk2021isaacgym,mittal2023orbit,mittal2025isaaclab}. Its benchmark objectives nevertheless inherited dense control costs: pose and velocity tracking are combined with torque, action, and smoothness regularizers \citep{peng2018deepmimic,hwangbo2019learning,lee2020learning,rudin2022learning,radosavovic2024realworld}.

Several lines of work have been proposed to reduce the reliance on hand-tuned dense rewards. Sparse and goal-conditioned methods reuse failures or schedule auxiliary objectives \citep{andrychowicz2017hindsight,riedmiller2018learning}; intrinsic motivation can guide a robot toward a binary task reward \citep{schwarke2023curiosity}; and episode-based black-box RL can exploit sparse or non-Markovian trajectory scores \citep{otto2023blackbox}.
A complementary line replaces hand-coded terms with semantic evaluators: language-model reward design, image-language rewards, video-based RoboCLIP, code-generating Eureka and its sim-to-real successor DrEureka, and zero-shot VLM reward models \citep{kwon2023reward,ma2023liv,sontakke2023roboclip,ma2024eureka,ma2024dreureka,rocamonde2024visionlanguage}.
Recent theory studies feedback available only for complete trajectories or trajectory segments \citep{zhang2025trajectory,du2025segment}, while recent VLA post-training demonstrates that long-horizon manipulation can be learned from a simplified binary trajectory-level outcome reward \citep{li2026simplevla}.

\section{Background}
\label{sec:problem_formulation}

We consider the following infinite-horizon optimal Markov decision process (MDP) objective, $J(\Theta) = \mathbb{E}_{s_t, a_t \sim \pi_\Theta} \Big[ \sum_{t = 0}^\infty \gamma^t R(s_t, a_t) \Big],\ s_0 \sim \rho_0$,
where $s_t \in \mathcal{S} \subset \mathbb{R}^{n_s}$ and $a_t \in \mathcal{A} \subset \mathbb{R}^{n_a}$ are continuous states and actions, $\Theta$ parameterizes the Gaussian policy $\pi_\Theta$, $R(\cdot,\cdot)$ is the reward function, $\gamma \in [0, 1)$ is the discount factor, and $\rho_0$ is the initial state distribution. Policy-gradient methods optimize the objective $J(\Theta) = \mathbb{E}_{\pi} \Big[ \widehat{A}_t \log \pi_\Theta(a_t|s_t) \Big]$,
where $\widehat{A}_t = \widehat G_t - V^{\pi}(s_t)$ is the advantage, $\widehat G_t$ is the bootstrapped return target estimated from a rollout, and $V^{\pi}(\cdot)$ is the value function of $\pi_\Theta$. PPO~\citep{schulman2017proximal} further employs ratio clipping to limit excessive policy updates over multiple gradient steps, yielding the surrogate objective $J^{\mathrm{PPO}}(\Theta)
    = \mathbb{E}_{\pi} \Big[\min \Big(\frac{\pi_\Theta(a_t|s_t)}{\pi_{\Theta_{\mathrm{old}}} (a_t|s_t)} \widehat{A}_t, \text{clip}(\frac{\pi_\Theta(a_t|s_t)}{\pi_{\Theta_{\mathrm{old}}}(a_t|s_t)}, 1-\epsilon, 1+\epsilon) \widehat{A}_t
  )\Big) \Big]$,
where $\epsilon$ is the clipping parameter. Trajectory optimization formulates the original optimal control problem as a constrained optimization problem:
\begin{equation}
    \max_{s_{0:T},a_{0:T-1}}
    \sum_{t=0}^{T} R(s_t,a_t),
    \quad\text{s.t.} \quad
    s_{t+1}=f(s_t,a_t),
    \quad (s_t,a_t)\in \mathcal{S} \times \mathcal{A}.
    \label{eq:direct_trajectory_optimization}
\end{equation}
which is solved by numerical optimizers such as IPOPT~\citep{ipopt}, where $f(\cdot, \cdot)$ is the dynamics. Unlike RL approaches that collect rollout states from the environment, trajectory optimization algorithms treat both states and actions as variables and enforce dynamics through equality constraints.

\section{Reward Shaping for Stabilization Control Problems}
\label{sec:method}

\begin{wrapfigure}[17]{r}{0.60\textwidth}
  \centering
  \setlength{\casadisweepheight}{0.352\linewidth}
  \begin{subfigure}[b]{0.516\linewidth}
    \centering
    \includegraphics[height=\casadisweepheight]{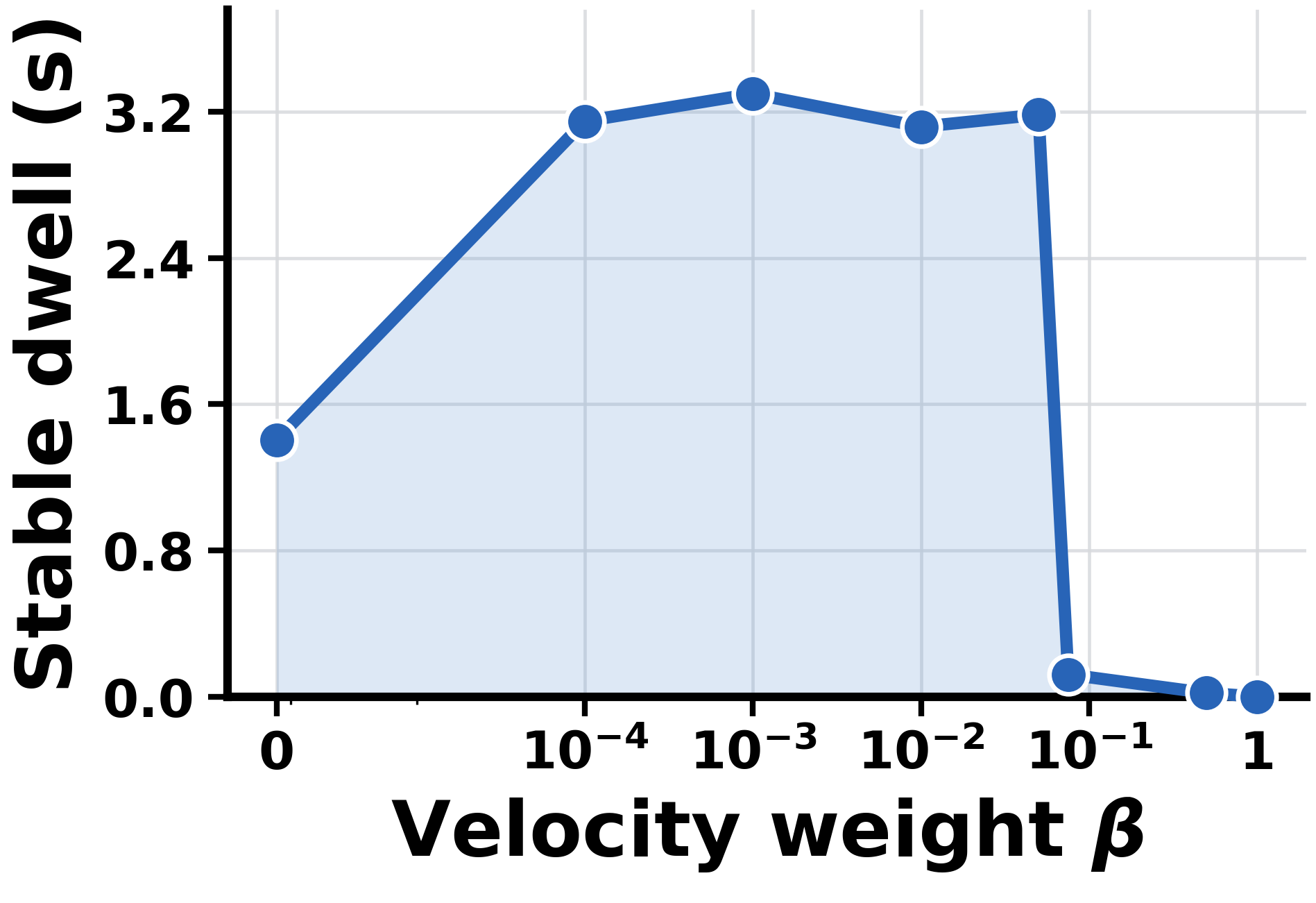}
    \caption{Acrobot}
    \label{fig:casadi_a}
  \end{subfigure}%
  \begin{subfigure}[b]{0.480\linewidth}
    \centering
    \includegraphics[height=\casadisweepheight]{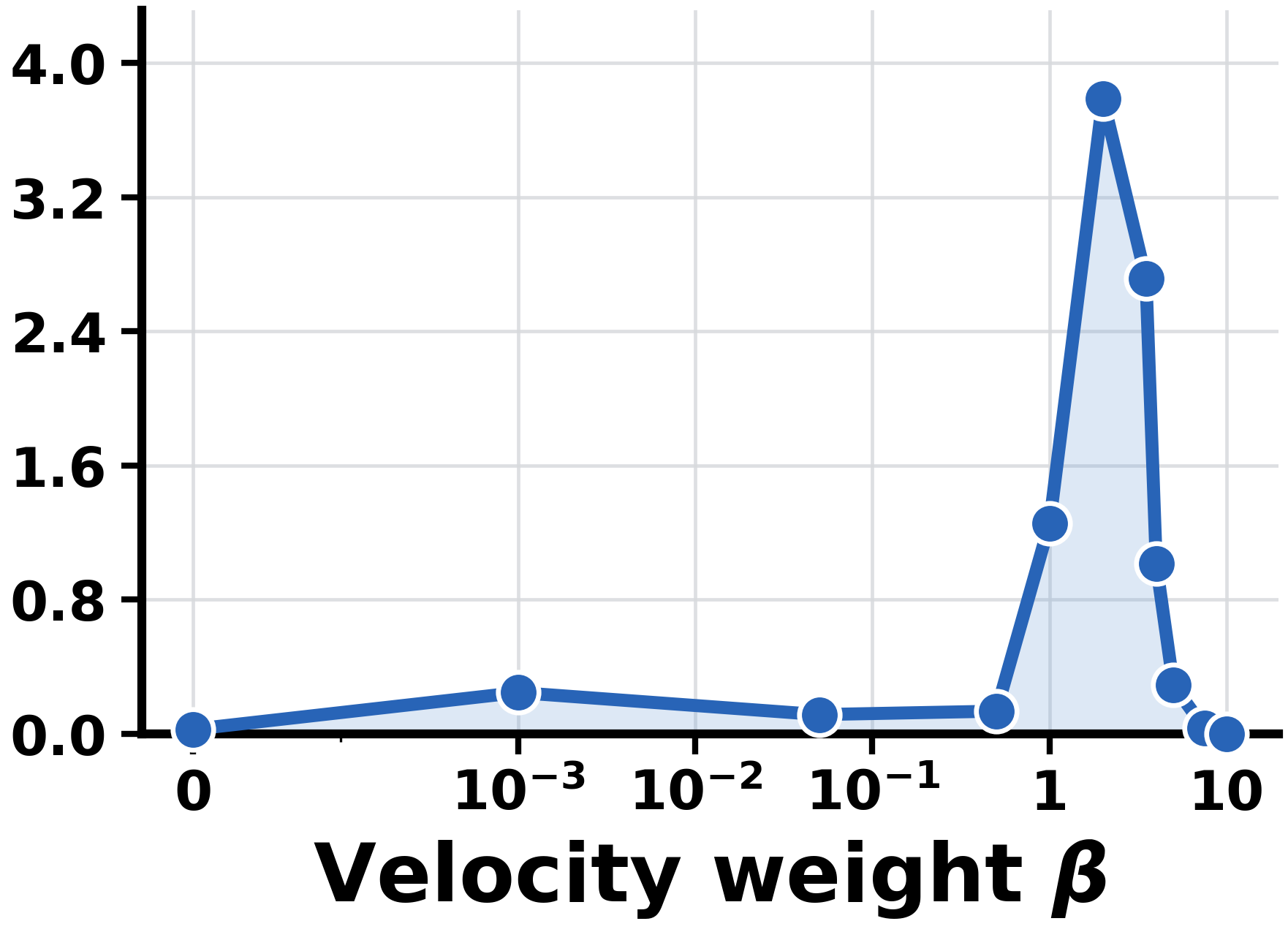}
    \caption{Pendubot}
    \label{fig:casadi_p}
  \end{subfigure}
  \caption{\textbf{A velocity penalty has task-dependent effects in direct trajectory optimization,} shown by the mean stable dwell for IPOPT-optimized Acrobot and Pendubot trajectories. Moderate velocity regularization helps in different ranges for the two plants, while an overly large weight suppresses successful motion in both.}
  \label{fig:casadi}
\end{wrapfigure}

Stabilization commonly uses the quadratic reward $R(s,a)=-(s^\top Qs+a^\top Ma)$, where $Q$ and $M$ weight state and control error respectively. The origin is the equilibrium. Direct trajectory-optimization methods often use positive-definite $Q$ and $M$ for numerical regularity.

\paragraph{Influence of velocity terms.}

\Cref{fig:casadi} illustrates that the numerical role of this regularizer is neither universal nor monotone. We directly optimize 600-step open-loop trajectories for Acrobot and Pendubot~\citep{spong} with IPOPT via CasADi~\citep{andersson2019casadi}, using a  quadratic reward $R_\beta^{\mathrm{TO}} =1-3\bigl(\theta_1^2+\theta_{2,a}^2\bigr) -\beta\bigl(\dot\theta_1^2+\dot\theta_2^2\bigr),$
where $\theta_1$ denotes the absolute angle of the first link, $\theta_2$ is the relative angle of the second link, and $\theta_{2,a}=\theta_1+\theta_2$ represents the absolute angle of the second link. Note that here the control weighting matrix is set to $M=0$, relying instead on explicit torque bounds $a \in [a_{\min}, a_{\max}]$ in direct transcription to prevent unbounded control actions.

For Acrobot, mean dwell rises from $1.41$ s at $\beta=0$ to roughly $3.2$--$3.4$ s over the moderate-weight regime, then collapses near zero once the velocity term becomes dominant. Pendubot instead has almost no dwell at $\beta=0$, reaches its best mean dwell of $3.8$ s near $\beta=2$, and again collapses for a sufficiently large coefficient. Thus, the same physically plausible term reshapes the objective differently across direct-optimization problems and can eventually overconstrain task progress.

This failure is caused by the second-order solver it uses. Let $H_f = \nabla^2 f(\mathbf{w})$ be the singular objective Hessian and $Z(\mathbf{w})$ be an orthonormal basis for the dynamic constraint nullspace $\ker(J)$, where $\mathbf{w}$ denotes the optimization variables and $J(\mathbf{w})$ is the constraint Jacobian. If $\ker(H_f)\cap\ker(J)$ is nonempty, then the reduced Hessian $Z^\top H_fZ$ is singular, and the solution set forms a non-unique manifold $\mathcal{W}^* = \mathbf{w}^* + \ker(H_f)\cap\ker(J)$, producing non-unique directions, KKT ill-conditioning, and solver instability. Appendix~\ref{app:hessian} gives the full derivation.

\paragraph{How RL achieves the control objective.}

The failure of a specific numerical solver does not mean that the control objective cannot be achieved without velocity rewards; we show next that it can, using a trajectory-level view of RL. A stochastic policy $\pi_\Theta(a\mid s)$ together with the transition kernel $p_{\mathrm{dyn}}$ induces a distribution over state-action trajectories $\bar\tau=(s_0,a_0, s_1, \ldots, a_{T-1}, s_T)$; marginalizing the action gives the policy-dependent closed-loop kernel
\begin{equation}
    \mathcal K_\Theta(s'\mid s)
    =
    \int_{\mathcal A}
    p_{\mathrm{dyn}}(s'\mid s,a)
    \pi_\Theta(a\mid s)
    \,da,
    \label{eq:closed_loop_kernel}
\end{equation}
so that the distribution over state trajectories $\tau_s=(s_0,\ldots,s_T)$ is given by $\mathcal P_\Theta(\tau_s)=\rho_0(s_0)\prod_{t=0}^{T-1}\mathcal K_\Theta(s_{t+1}\mid s_t)$, and policy optimization induces the chain $\boxed{\pi_\Theta \longrightarrow \mathcal K_\Theta \longrightarrow \mathcal P_\Theta(\tau_s)}$.
Accordingly, RL can be viewed as an indirect optimization over the policy-realizable trajectory-distribution family $\mathbb P_\Pi = \left\{\mathcal P_\Theta(\tau_s): \pi_\Theta\in\Pi \right\}$.
Unlike the direct transcription of \Cref{eq:direct_trajectory_optimization}, which manipulates a single trajectory under explicit dynamics constraints, PPO only reshapes $\mathcal P_\Theta(\tau_s)$ indirectly through the actor, Appendix~\ref{app:trajectory_level_view} expands this comparison.

We now separate the information used to evaluate a trajectory from the information used to generate it. Let the instantaneous reward depend only on the zeroth-order state, $R(s_t, a_t) = r_0(q_t)$, where $s_t = (q_t, v_t)$ is the full state and $v_t = \dot q_t$ is the first-order state. Let $\mathcal R_0(\tau_s)=\sum_{t=0}^{T}\gamma^t r_0(q_t)$ be the zeroth-order trajectory reward, so the objective is
\begin{equation}
    J(\Theta)
    =
    \mathbb E_{\tau_s\sim \mathcal P_\Theta}[\mathcal R_0(\tau_s)]
    =
    \int \mathcal P_\Theta(\tau_s)\mathcal R_0(\tau_s)\,d\tau_s.
    \label{eq:trajectory_objective}
\end{equation}

Every zeroth-order term used in this paper is a bounded function of configuration error that is uniquely maximized at the target $q^\ast$: writing $r^\ast=r_0(q^\ast)$, for every $\delta>0$ there is a gap $\epsilon_0(\delta)>0$ such that $r_0(q)\;\le\;r^\ast-\epsilon_0(\delta),\ \forall q\notin\mathcal N_\delta(q^\ast):=\{q:\lVert q-q^\ast\rVert<\delta\}$.
For trajectories $\tau_s^+,\tau_s^-$ that agree outside a common index set $S\subseteq\{0,\ldots,T\}$, with $q_t^+\in\mathcal N_\delta(q^\ast)$ (dwelling near the target) and $q_t^-\notin\mathcal N_\delta(q^\ast)$ (away from it, e.g.\ overshooting or oscillating) for every $t\in S$, this gap propagates directly to the return because the two trajectories agree termwise outside $S$:
\begin{equation}
    \mathcal R_0(\tau_s^+)-\mathcal R_0(\tau_s^-)
    =
    \sum_{t\in S}\gamma^t\bigl(r_0(q_t^+)-r_0(q_t^-)\bigr)
    \;\ge\;
    \epsilon_0(\delta)\sum_{t\in S}\gamma^t
    \;>\;0.
    \label{eq:dwell_return_gap}
\end{equation}
So $\mathcal R_0$ strictly prefers the dwelling trajectory without first-order term $\dot q_t$ appearing anywhere in $r_0$: shifting probability mass $\varepsilon$ from the oscillating class to the dwelling class strictly increases $J$ by at least $\varepsilon\epsilon_0(\delta)\sum_{t\in S}\gamma^t$.

The same holds at the level of the PPO advantage rather than the trajectory distribution. At current policy, the clipped surrogate has the standard first-order policy gradient \citep{schulman2017proximal},
\begin{align}
    \left.\nabla_\Theta J^{\mathrm{PPO}}(\Theta)\right|_{\Theta_{\mathrm{old}}}
    &=
    \mathbb E_\pi\left[A^\pi(q,v,a)\nabla_\Theta\log\pi_\Theta(a\mid q,v)\right],
    \\
    Q^\pi(q,v,a)
    &=
    r_0(q)+\gamma\mathbb E_{s'\sim p_\mathrm{dyn} (\cdot \mid q,v,a)} [V^\pi(s')].
    \label{eq:zeroth_reward_policy_gradient}
\end{align}

Since $r_0(q)$ is identical across actions at a fixed $(q,v)$, it cancels exactly from the advantage $A^\pi=Q^\pi-V^\pi$: the advantage is determined solely by how the action $a$ perturbs the distribution of the future zeroth-order return through $V^\pi(q,v)=\mathbb E_{a\sim\pi(\cdot\mid q,v)}[Q^\pi(q,v,a)]$ (Appendix~\ref{app:trajectory_level_view} derives this cancellation explicitly).
Crucially, $V^\pi$ is a function of $v$ even though $r_0$ is not, because $v$ enters the dynamics $p_{\mathrm{dyn}}$ and hence shapes the distribution of future configurations $q_{t+1:T}$. First-order reward terms are therefore unnecessary for providing a braking or stabilization learning signal; this statement concerns trajectory evaluation only, and whether the policy can realize the preferred trajectory depends separately on its observation.

\paragraph{Velocity observations are necessary.}

Although velocities need not appear in the reward, they remain critical in the observation. We represent first-order information by the momentum $p\in\mathbb R^{n_s}$. Consider the mechanical system with the canonical phase-space state vector $x = [q, p] ^ \top \in \mathbb{R}^{2n_s}$, where $q\in\mathbb R^{n_s}$ is the generalized configuration, $p=M(q)v$ is the generalized momentum, and $a\in\mathbb R^{n_a}$ is the control input. Its dynamics are
\begin{equation}
    \dot q=\nabla_p H_0(q,p),
    \qquad
    \dot p=-\nabla_q H_0(q,p)+B(q)a,
\end{equation}
where $H_0(q,p)=\frac12p^\top M(q)^{-1}p+U(q)$. We make the following assumption:
\begin{assumption}
\label{assumption}
    (policy-supplied dissipation) In a neighborhood of the target $x^\ast=[ q ^ \ast, 0]^\top$, the dynamics seen from the policy action interface contain no velocity-dependent dissipative force outside the policy. We also assume that $x^\ast$ is locally stabilizable by some admissible full-state feedback $a=g_{qp}(q,p)$.
\end{assumption}

This assumption excludes intrinsically unstabilizable tasks but does not require full actuation. At deterministic deployment, a continuously differentiable zeroth-order policy has the form $a=g_q(q)$. Its closed-loop vector field is $l_q=(\nabla_pH_0,-\nabla_qH_0+B g_q)$, whose divergence in canonical phase space is
\begin{equation}
    \nabla_{q,p}\cdot l_q
    =
    \sum_i\frac{\partial^2H_0}{\partial q_i\partial p_i}
    -
    \sum_i\frac{\partial^2H_0}{\partial p_i\partial q_i}
    +
    \nabla_p\cdot[B(q)g_q(q)]
    =0.
    \label{eq:q_only_divergence}
\end{equation}
Hence, position-only feedback can reshape the force field but cannot contract phase-space volume. Indeed, we have the following theorem:
\begin{theorem}
\label{th:prop1}
    Under Assumption~\ref{assumption}, no continuously differentiable memoryless policy $a=g_q(q)$ can make $x^\ast$ locally asymptotically stable.
\end{theorem}

The proof of Theorem~\ref{th:prop1} is included in Appendix~\ref{app:proof}; a double-integrator example in Appendix~\ref{app:illustr} makes the associated observation aliasing explicit.
The argument is independent of the rank and dimension of $B(q)$, so it applies to both fully actuated and underactuated systems whenever the full-state stabilizability premise holds. It rules out asymptotic attraction, not transient target visits: a position-only policy may enter the target region frequently while failing to remain there.
Conversely, passive damping, lower-level PD control, or dissipative contact that violates Assumption~\ref{assumption} may permit long stable dwell without exposing velocity to the learned policy. The required velocity information must be available to the component that supplies dissipation, but that component need not be the actor.

In RL context, an undiscounted infinite-horizon cost-to-go can serve as a control Lyapunov function subject to suitable stabilizability and regularity conditions; under the reward convention, the candidate is $-V^\pi$ \citep{camilli2008control,kamalapurkar2018reinforcement}. A discounted value function is not automatically a Lyapunov function and requires additional dominance condition \citep{gaitsgory2015stabilization}. Likewise, closed-loop stability with an approximate critic requires explicit approximation-error and learning conditions \citep{kamalapurkar2018reinforcement}. In either case, the critic cannot remove the structural obstruction imposed on a zeroth-order actor.

\section{Experiments}
\label{sec:isaaclab_asymmetry}

In this section, we validate our analysis on seven Isaac Lab control
problems spanning underactuated swing-up (Acrobot and Pendubot), cart-based swing-up (Cartpole and Double Cart Pendulum), floating-base flight (Quadcopter), articulated manipulation (Franka Reach), and contact-rich whole-body motion (Humanoid). Their varied dimensions, actuation, and dissipation test the conclusion beyond variants of a single plant.

\paragraph{Experiment setup.}

For each task we write the scalar reward as $r_\beta(q_t,v_t)=r_0(q_t)-\beta e_v(v_t)$, where $\beta\ge 0$ and $q$ only includes configuration-derived task variables such as Cartesian position and orientation.
Let $e_c(\theta)=1-\cos \theta$, let $d_p$ denote position error, and let $e_R=1-(q^\top q^*)^2$ denote the orientation error (quaternion-dot form). For the two-link systems, $\theta_{2,a}=\theta_1+\theta_2$ is the absolute second-link angle. \Cref{tab:isaaclab_reward_family} gives the complete reward design for all tasks. For Humanoid, $\sigma$ is the logistic function,
$b(d_{xy})=\mathbbm{1}[d_{xy}\leq0.55] \operatorname{clip}((0.55-d_{xy})/0.20,0,1)$ gates braking near the target, and $e_{\mathrm{stand}}$ is a zeroth-order pseudo-Huber combination of root orientation, calibrated root height, joint-mean error, and joint-maximum error.
We evaluate every policy using a dynamical success set $\mathcal S_{\mathrm{succ}} = \{(q,v):e_q^{\mathrm{eval}}(q)\leq\epsilon_q, \ e_v^{\mathrm{eval}}(v)\leq\epsilon_v\}$, so removing velocity from a training reward never weakens the evaluation criterion. For trajectory $i$, define $C_{i,t}=\mathbbm{1}[(q_{i,t},v_{i,t})\in\mathcal S_{\mathrm{succ}}]$ and let $\mathfrak I_i$ be its set of contiguous successful index intervals. We report
\begin{equation}
  \mathrm{Success Rate}
  =\frac1N\sum_{i=1}^N\mathbbm{1}[\exists t:C_{i,t}=1],
  \qquad
  \mathrm{Dwell}
  =\frac{\Delta t}{N}\sum_{i=1}^N
    \max_{I\in\mathfrak I_i}|I|,
  \label{eq:evaluation_metrics}
\end{equation}
Thus, success is the fraction of trajectories that ever enter the success set, whereas dwell averages each trajectory's longest uninterrupted visit.
Exact predicates and rollout banks are in Appendix~\ref{app:implement} (\Cref{tab:success_thresholds}), PPO settings are in Appendix~\ref{app:ppo_hyperparameters}, and optimization traces are in Appendix~\ref{app:additional}.

\begin{table*}[t]
\centering
\caption{Reward family used in the Isaac Lab sweep. Every row has the form
$r_\beta=r_0-\beta e_v$. The $\beta=0$ condition removes velocity only from
the reward; the policy still receives the full task-defined configuration and
velocity observation.}
\label{tab:isaaclab_reward_family}
\footnotesize
\setlength{\tabcolsep}{4pt}

\begin{tabular}{p{0.20\textwidth} | p{0.50\textwidth} | p{0.25\textwidth}}

\Xhline{1.2pt}

Task & Zeroth-order term $r_0(q)$ & Velocity error $e_v(v)$ \\

\Xhline{1.2pt}

Acrobot & $1-3[e_c(\theta_1)+e_c(\theta_{2,a})]$
& $\dot\theta_1^2+\dot\theta_2^2$\\

Pendubot & $1-3[e_c(\theta_1)+e_c(\theta_{2,a})]$
& $\dot\theta_1^2+\dot\theta_2^2$\\

Cartpole & $1-3e_c(\theta)$
& $v_c^2+\dot\theta^2$ \\

Double cart pendulum & $2-3[e_c(\theta_1)+e_c(\theta_{2,a})]$
& $v_c^2+\dot\theta_1^2+\dot\theta_2^2$\\

Quadcopter & $1-3\tanh(d_p/0.8)-e_R$, $e_R = 1 - (q ^ \top q ^ \ast) ^ 2$
& $\lVert v_{\mathrm{body}}\rVert_2^2$\\

Franka reach & $1-3\tanh(d_p/0.8)-3e_q$, $e_q=\sqrt{\operatorname{mean}_j(q_j-q_j^*)^2}$
& $\lVert\dot q\rVert_2^2$\\

Humanoid & $r_{\mathrm{default,ZO}}-\alpha(d_{xy})\tanh(d_{xy}/s_p)
+2\sigma((0.35-d_{xy})/0.04)-3b(d_{xy})e_{\mathrm{stand}}$
& $\lVert v_{\mathrm{root}}\rVert_2^2+
\lVert s_\omega\omega_{\mathrm{root}}\rVert_2^2+
\sum\lVert s_q\dot q\rVert_2^2$\\

\Xhline{1.2pt}

\end{tabular}
\end{table*}

\paragraph{Velocity rewards are not required for successful control.}

\begin{figure*}[t]
    \centering
    \setlength{\betasweepheight}{0.181\textwidth}
    \begin{subfigure}[b]{0.263\textwidth}
        \centering
        \includegraphics[height=\betasweepheight]{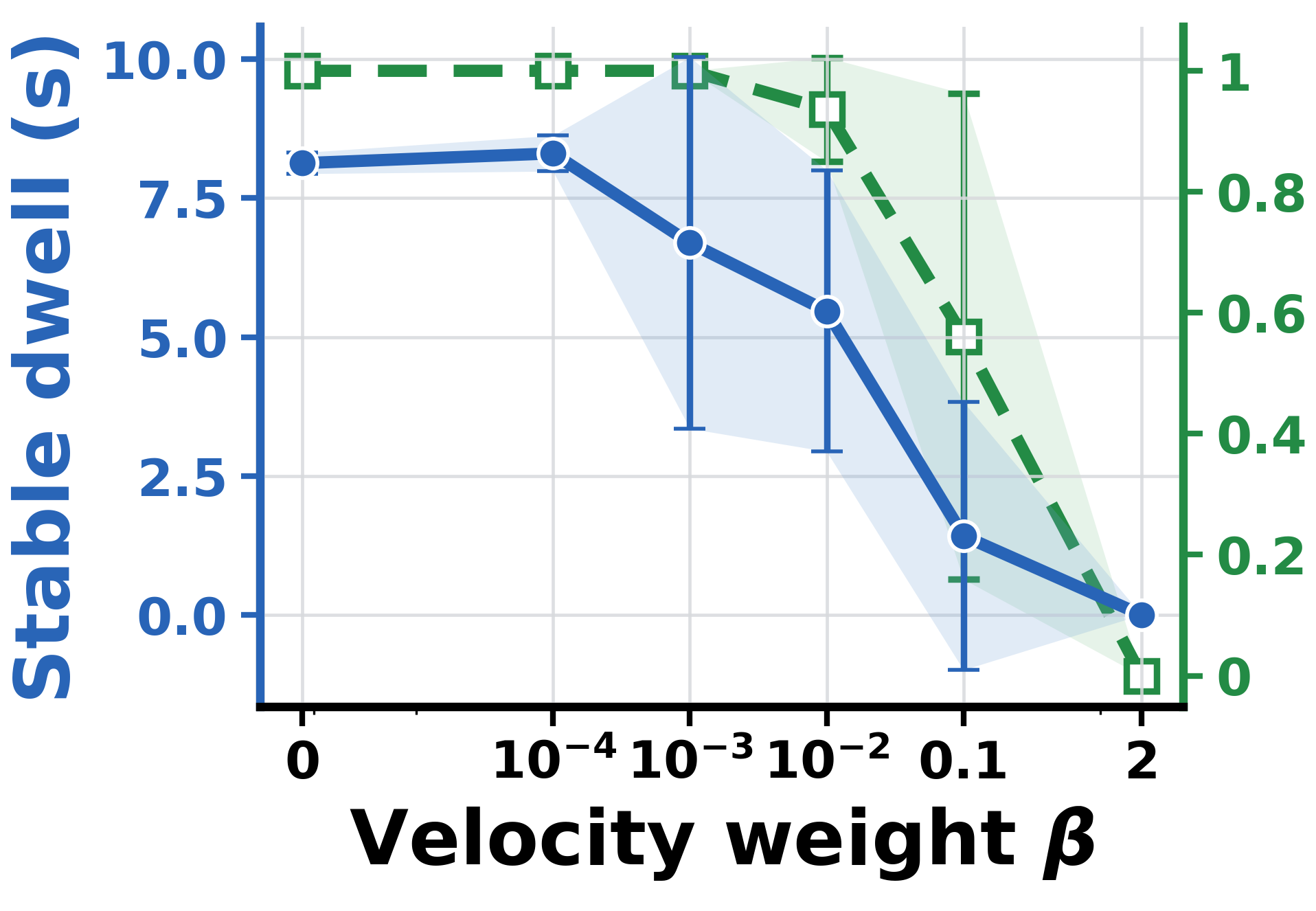}
        \caption{Acrobot}
        \label{fig:beta_sweep_a}
    \end{subfigure}%
    \begin{subfigure}[b]{0.238\textwidth}
        \centering
        \includegraphics[height=\betasweepheight]{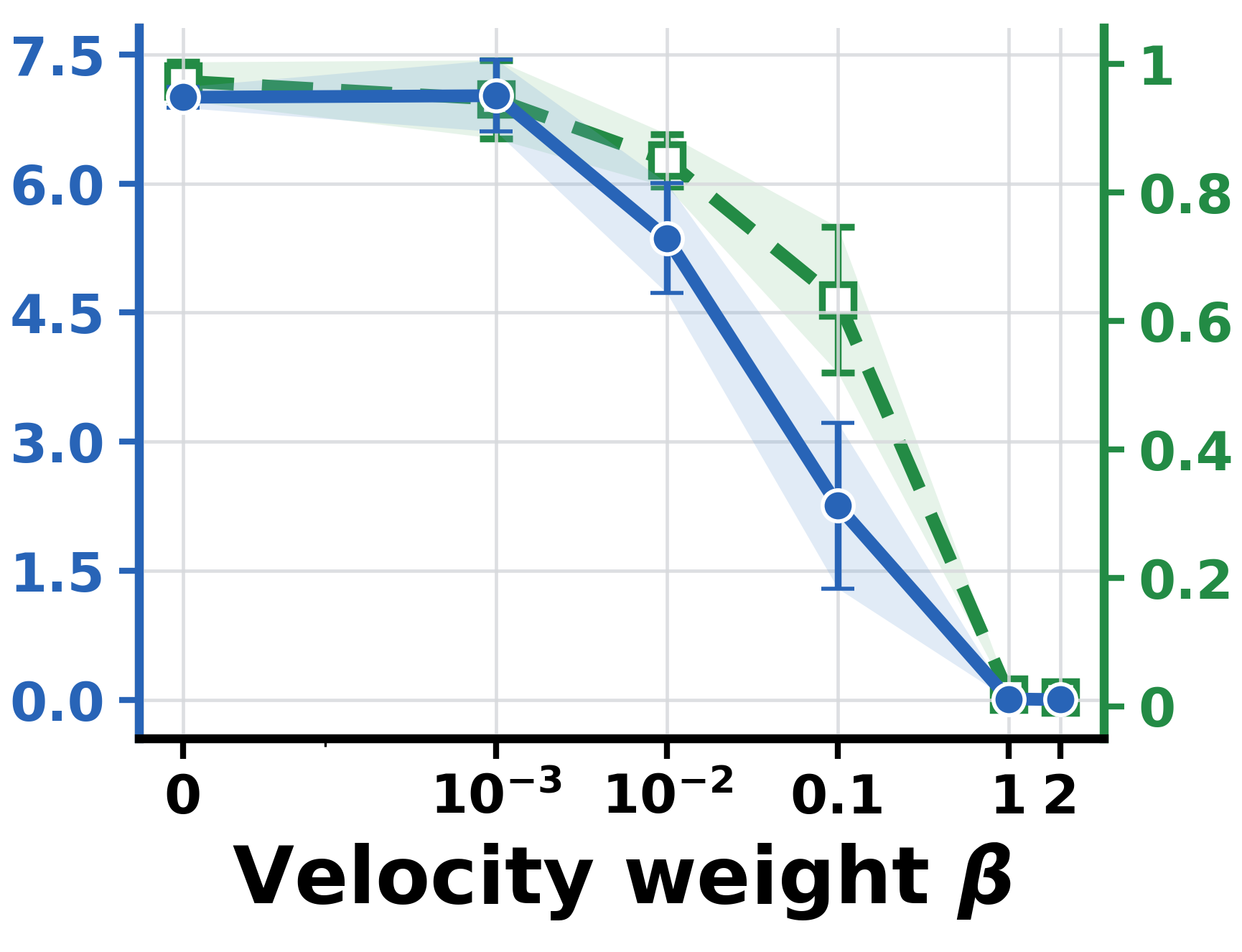}
        \caption{Pendubot}
        \label{fig:beta_sweep_p}
    \end{subfigure}%
    \begin{subfigure}[b]{0.245\textwidth}
        \centering
        \includegraphics[height=\betasweepheight]{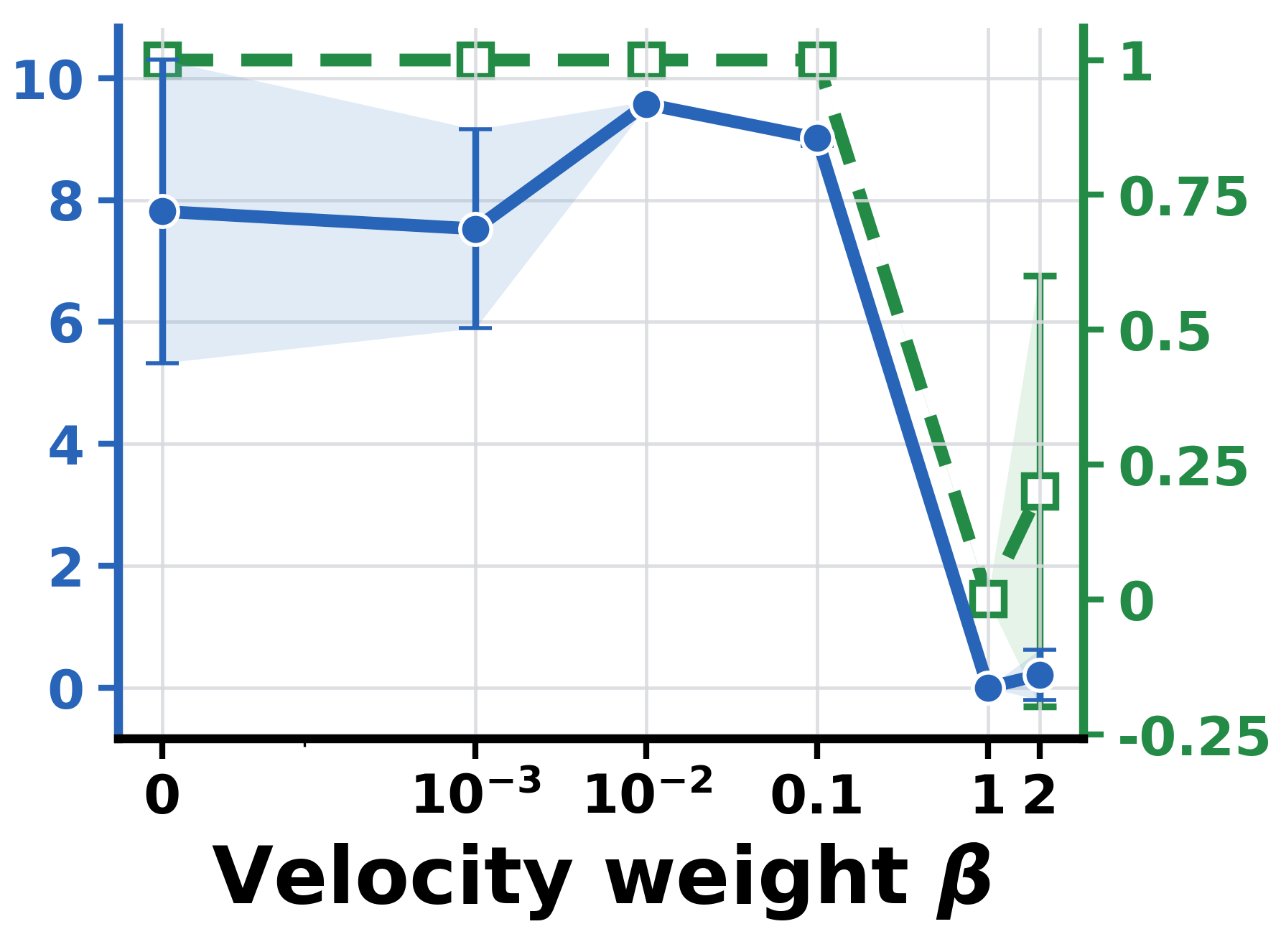}
        \caption{Cartpole}
        \label{fig:beta_sweep_c}
    \end{subfigure}%
    \begin{subfigure}[b]{0.252\textwidth}
        \centering
        \includegraphics[height=\betasweepheight]{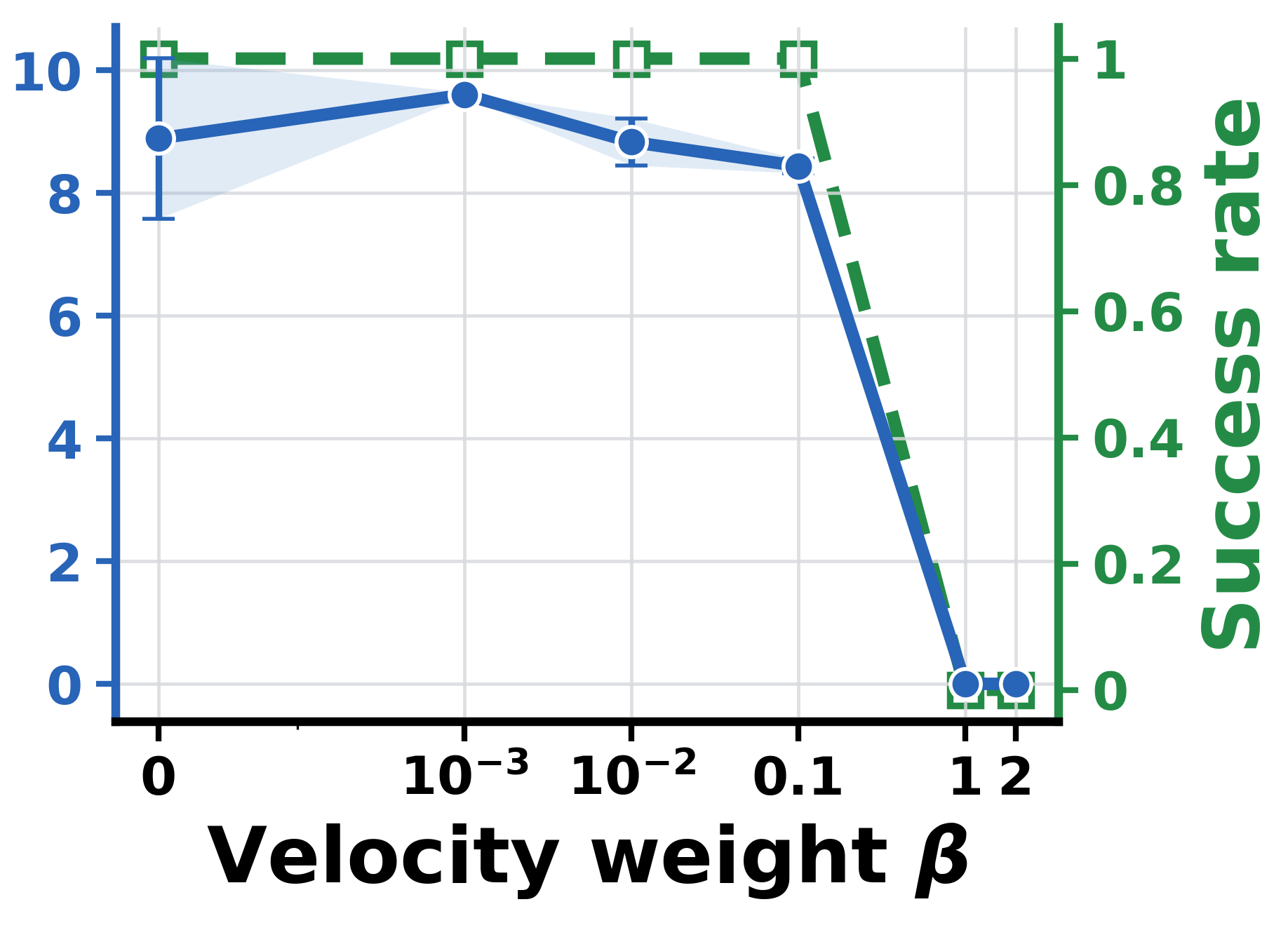}
        \caption{Double Cart Pendulum}
        \label{fig:beta_sweep_d}
    \end{subfigure}
    
    \begin{subfigure}[b]{0.270\textwidth}
        \centering
        \includegraphics[height=\betasweepheight]{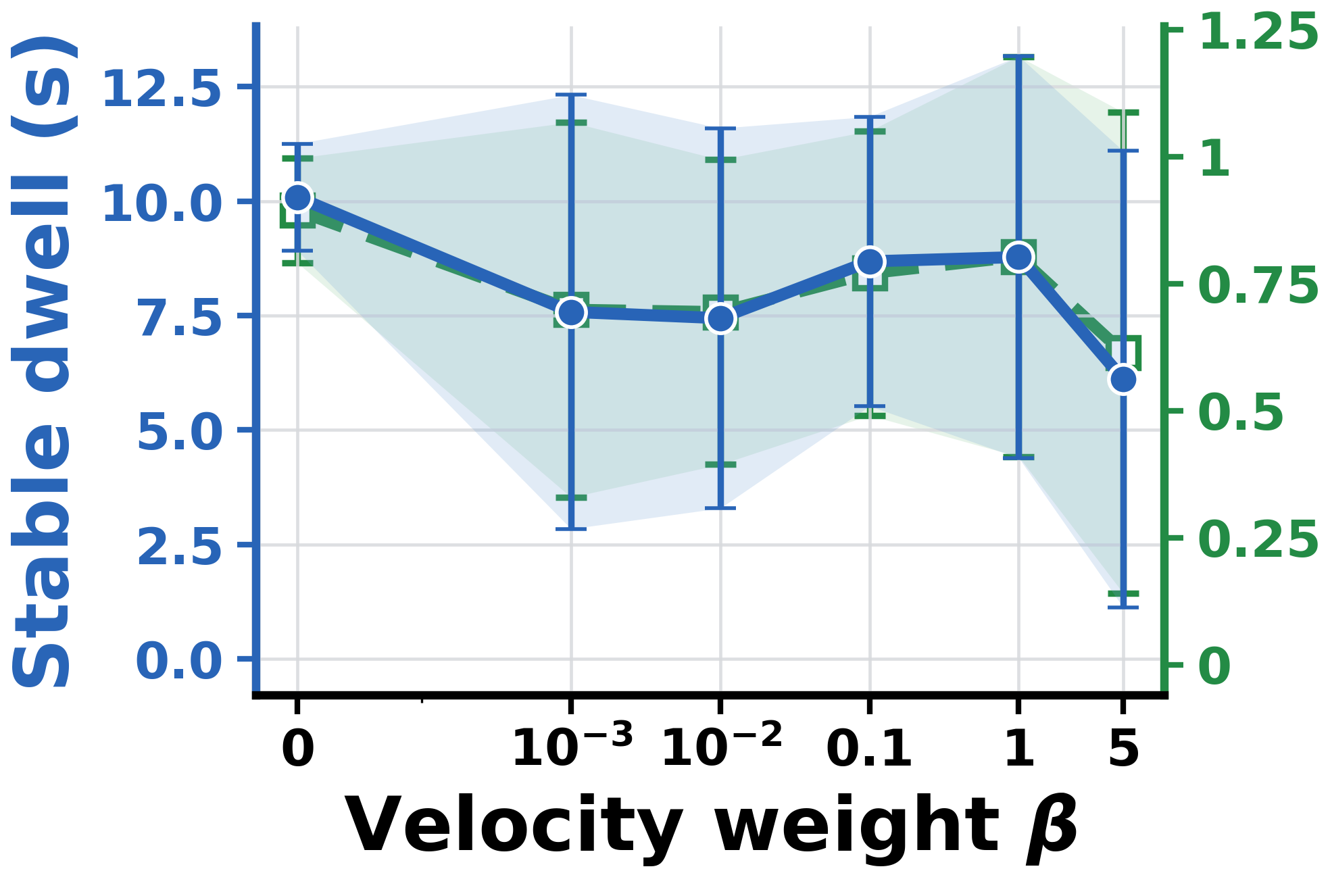}
        \caption{Quadcopter}
        \label{fig:beta_sweep_q}
    \end{subfigure}%
    \begin{subfigure}[b]{0.241\textwidth}
        \centering
        \includegraphics[height=\betasweepheight]{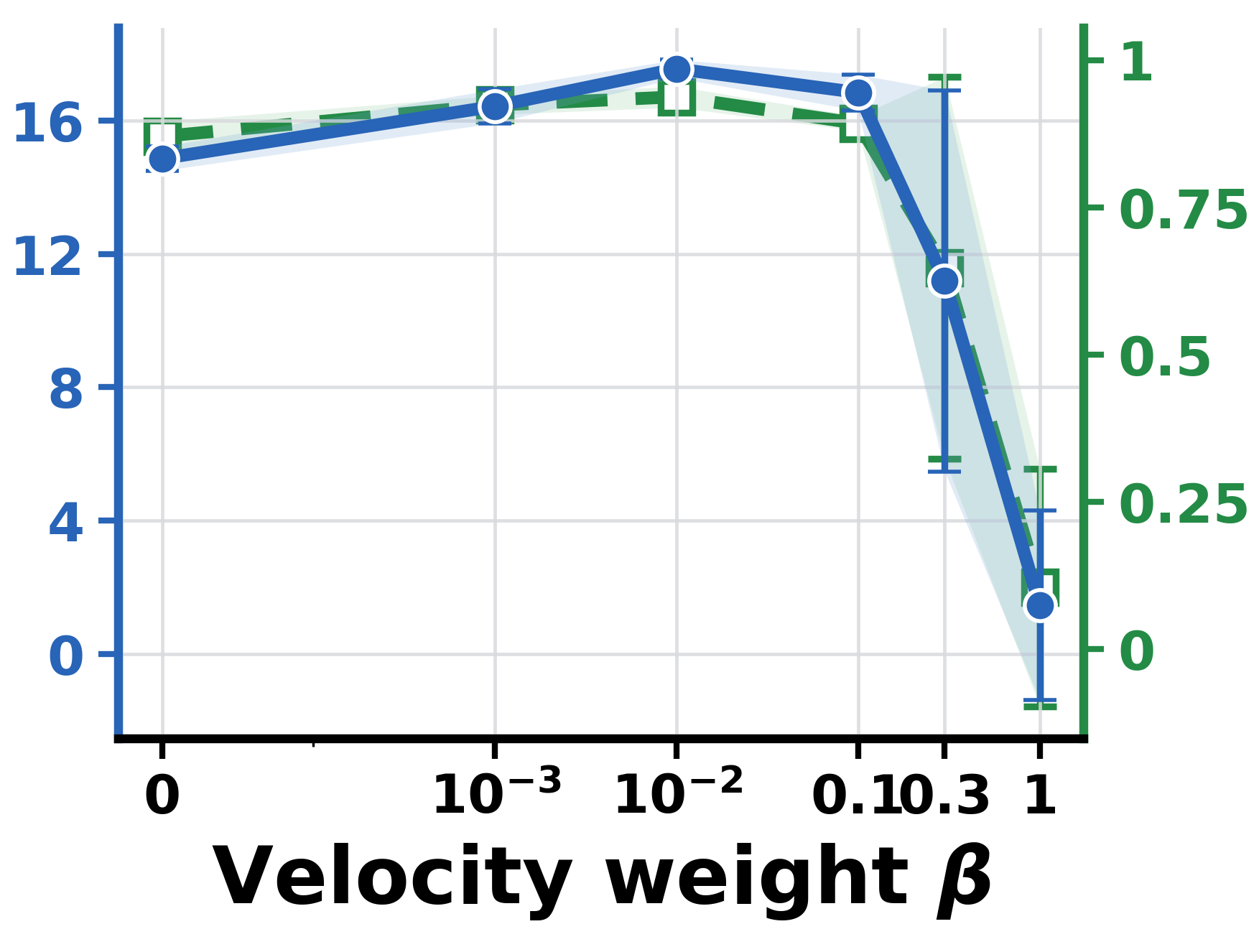}
        \caption{Franka Reach}
        \label{fig:beta_sweep_f}
    \end{subfigure}%
    \begin{subfigure}[b]{0.245\textwidth}
        \centering
        \includegraphics[height=\betasweepheight]{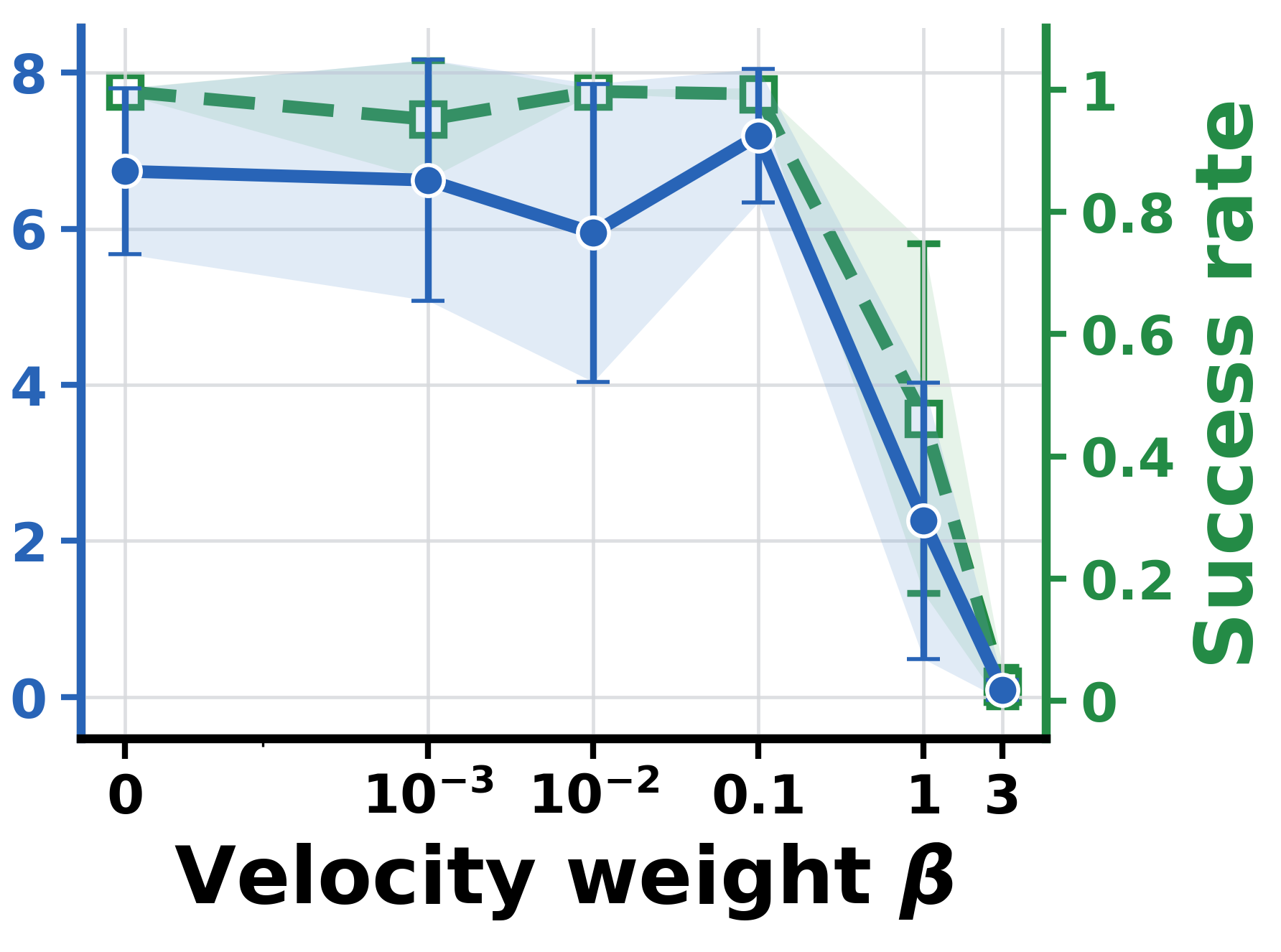}
        \caption{Humanoid}
        \label{fig:beta_sweep_h}
    \end{subfigure}

    \caption{Velocity-reward sweeps with full configuration and velocity observations. Blue solid curves show stable dwell and green dashed curves show success rate; bands are mean $\pm$ one population standard deviation over five seeds. Every task learns reaching, braking, and sustained stabilization at $\beta=0$, while sufficiently large velocity penalties generally reduce both metrics.}
    \label{fig:beta_sweep}
\end{figure*}

With full observations, removing the explicit velocity term does not remove the learning signal for braking. As shown in \Cref{fig:beta_sweep} at $\beta=0$, success rates are $1.00$, $0.973$, $1.00$, $1.00$, $0.894$, $0.870$, and $0.997$ for Acrobot, Pendubot, Cartpole, Double Cart Pendulum, Quadcopter, Franka, and Humanoid, respectively.
Their corresponding mean stable dwells are $8.13$, $7.00$, $7.81$, $8.88$, $10.09$, $14.86$, and $6.74$ seconds.  Because each success test includes a velocity threshold, these are not merely target visits: the policies approach, brake, and remain inside a dynamically defined success set for sustained periods.

\textbf{This finding is the empirical counterpart of \Cref{eq:zeroth_reward_policy_gradient}.} The instantaneous reward need not observe velocity to prefer braking actions: actions change future states, future configurations change $\mathcal R_0(\tau_s)$, and the critic propagates that
trajectory-level difference into the advantage. Full $(q,v)$ observations then let the actor assign different actions to the same configuration reached with different velocities.

\textbf{Larger $\beta$ is not uniformly beneficial.} For example, at the largest tested weight, Acrobot and Double Cart Pendulum have zero dwell and zero success; Franka falls from $14.86$ s to $1.46$ s; and Humanoid falls from $6.74$ s to $0.089$ s, demonstrating a clear cross-task pattern: an explicit velocity penalty is unnecessary at $\beta=0$, and making it dominant can sharply impair learning.

\paragraph{Velocity observations remain necessary.}

\begin{wrapfigure}[18]{r}{0.60\textwidth}
    \centering
    \includegraphics[width=\linewidth]{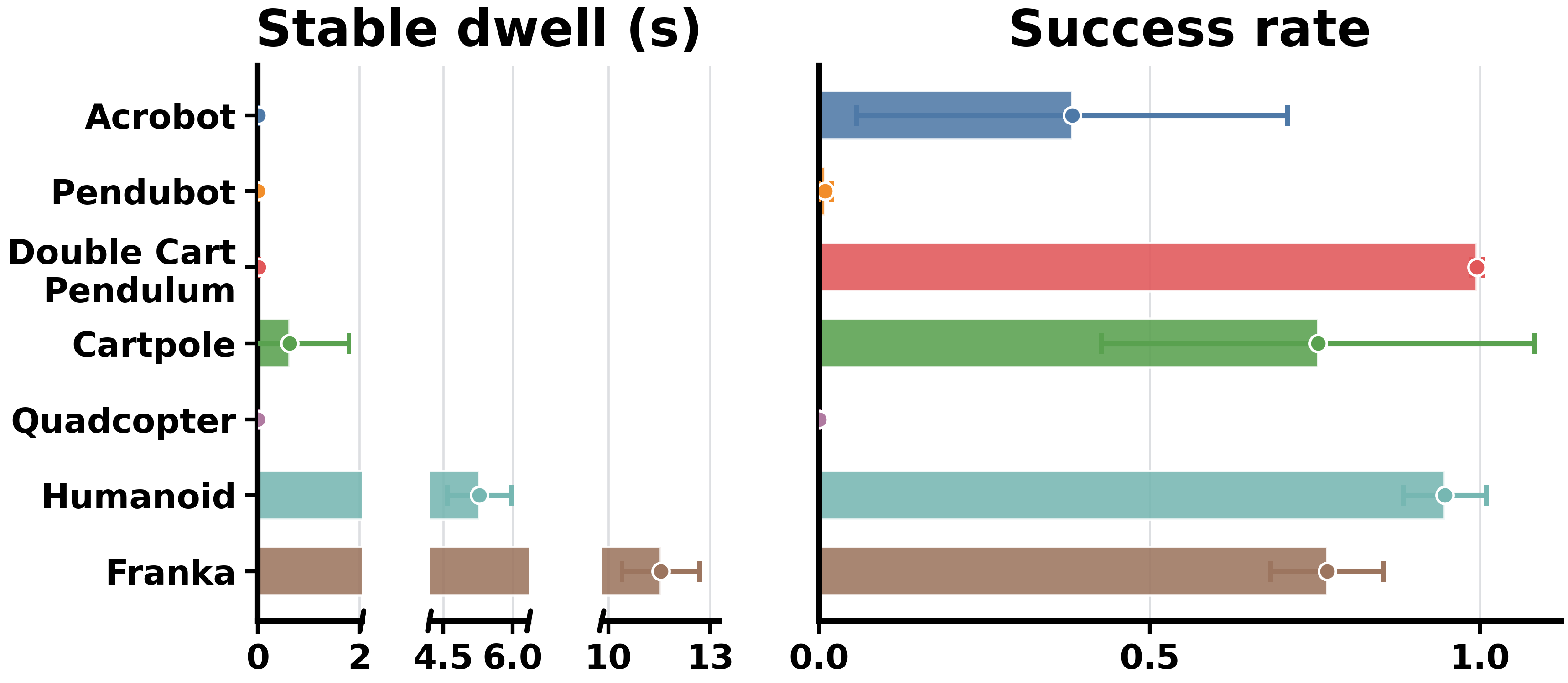}
    \caption{No-velocity-observation ablation at $\beta=0$. Each task keeps
    exactly the same zeroth-order reward and velocity-aware success predicate
    used in the full-observation experiment. Bars show the five-seed mean and
    population standard deviation of best dwell (left) and corresponding
    success rate (right).}
    \label{fig:no_obs}
\end{wrapfigure}

We hold $r_0$ fixed and remove only velocity from the policy observation. This leaves the trajectory ranking unchanged but restricts the feedback laws available to the actor. The resulting failure modes \textbf{separate transient reachability from sustained stabilization}.
As shown in \Cref{fig:no_obs}, Pendubot collapses from $7.00$ s dwell and $0.973$ success rate with full observations to $2.34\times10^{-4}$ s and $0.009$; Quadcopter reaches zero on both metrics. Double Cart Pendulum provides the clearest pass-through signature: success-ever remains $0.995$, yet dwell falls from $8.88$ s to $0.0166$ s. Cartpole behaves similarly but less extremely, retaining $0.755$ success while dwell drops from $7.81$ s to $0.63$ s. These policies can visit an upright configuration, but without velocity-conditioned braking they oscillate through it or fail to remain there.

The mixed high-dimensional results also expose \textbf{the boundary of the theoretical claim}. Franka and Humanoid degrade more modestly, from $14.86$ to $11.54$ s and from $6.74$ to $5.28$ s, respectively. Franka's lower-level joint-position servo and Humanoid's dissipative contact dynamics can supply contraction outside the learned memoryless actor. These mechanisms violate Assumption~\ref{assumption} and are therefore consistent with Theorem~\ref{th:prop1}.
Together, these results show that \textbf{reward sufficiency and observation sufficiency are distinct experimental properties}.

\paragraph{Why does reward clipping help?}

Our analysis also explains why reward clipping is an effective design heuristic. Using the zeroth-order terms $r_0$ and first-order errors $e_v$ from \Cref{tab:isaaclab_reward_family}, the clipping interventions for Pendubot and Franka leave $r_0$ unchanged and caps only $e_v$:
\begin{equation}
    r_{\beta,c}^{\mathrm P}=r_0^{\mathrm P}-\beta\min\left\{\dot\theta_1^2+\dot\theta_2^2,c\right\},\qquad
    r_{\beta,c}^{\mathrm F}=r_0^{\mathrm F}-\beta\min\left\{\lVert\dot q\rVert_2^2,c\right\},\qquad
    c=4.
    \label{eq:main_velocity_cap}
\end{equation}
To diagnose the learning distribution, we additionally use one-iteration probes under seed 42. Each probe independently collects one stochastic on-policy rollout at a saved checkpoint and then performs one PPO update, exposing its return targets, advantages, and critic loss.
Appendix~\ref{app:implement} gives the complete task-specific rewards and implementation details.

\begin{figure*}[h]
  \centering
    \begin{subfigure}[b]{0.49\textwidth}
        \centering
        \includegraphics[width=\textwidth]{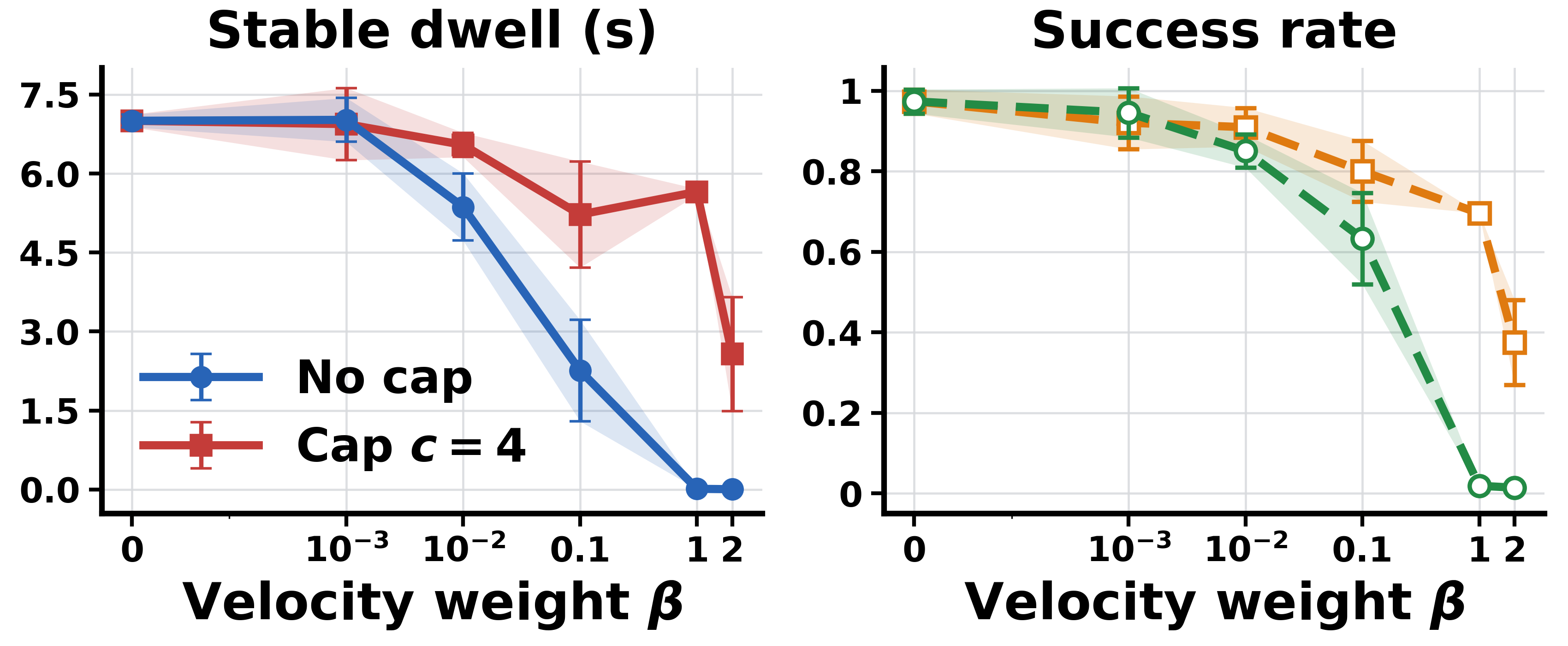}
        \caption{Pendubot}
    \end{subfigure}
    \hfill
    \begin{subfigure}[b]{0.49\textwidth}
        \centering
        \includegraphics[width=\textwidth]{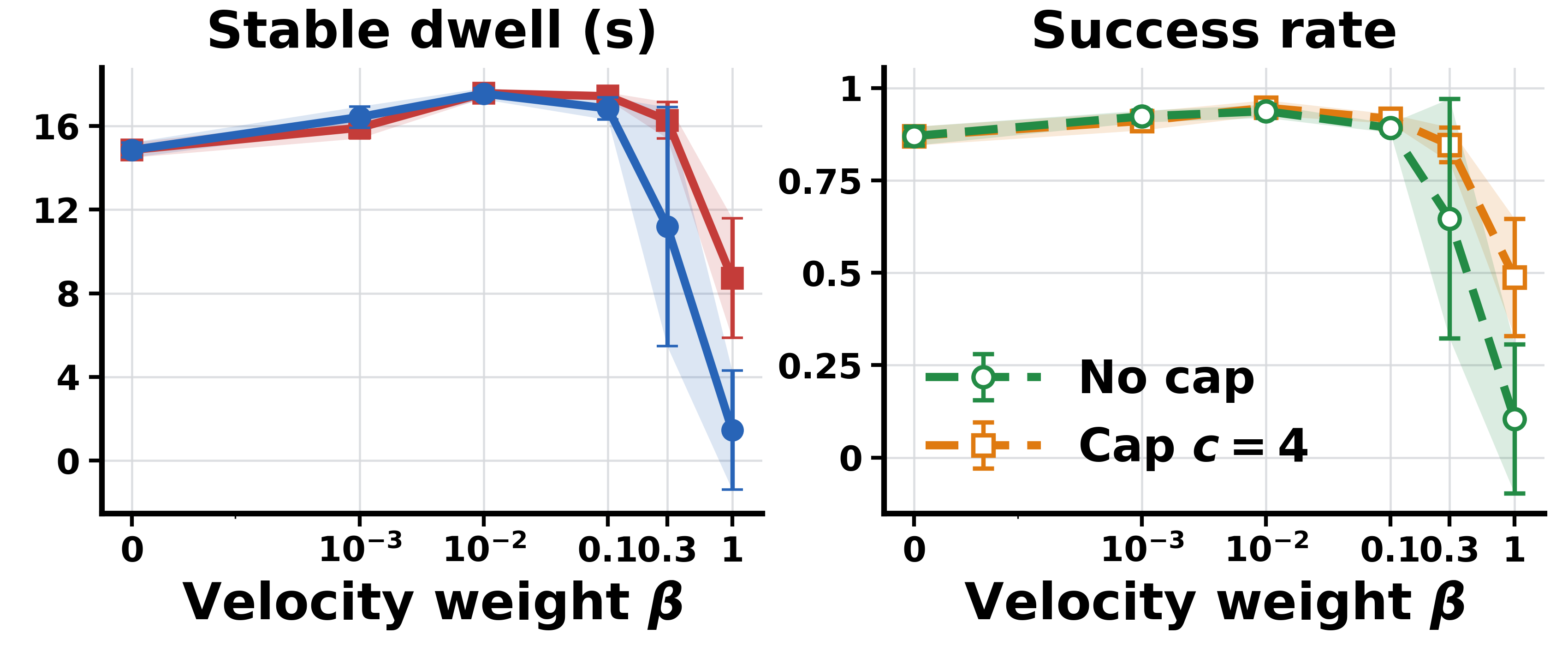}
        \caption{Franka Reach}
    \end{subfigure}
  \caption{\textbf{Capping only the squared-velocity error prevents the large-}$\boldsymbol{\beta}$ \textbf{performance collapse.} Five-seed beta sweeps for Pendubot (left) and Franka (right) compare the uncapped reward with the velocity cap $c=4$ in ~\Cref{eq:main_velocity_cap}. Curves and bands are the mean $\pm$ one population standard deviation. The cap leaves the successful low-$\beta$ regime intact and restores substantial performance where the raw reward fails.}
  \label{fig:clipping_performance}
\end{figure*}

\begin{table*}[h!]
  \centering
  \small
  \setlength{\tabcolsep}{7pt}
  \caption{\textbf{Reward clipping contracts return tails and critic error.} Dwell is the best evaluation duration in each seed-42 run. The remaining columns are means over the final five matched one-iteration probes; $\widehat G_t$ denotes the scalar return target at time step $t$.}
  \label{tab:clipping_probe}
  \begin{tabular}{llclrrr}
    \toprule
    Task & $\beta$ & Reward & Dwell (s) $\uparrow$
      & $\overline{\widehat G_t}$
      & $\overline{|\min_t\widehat G_t|}$
      & Value MSE $\downarrow$ \\
    \midrule
    Pendubot & 0 & Raw          & 6.87  & 31.9  & $9.10 \times 10^2$ & $3.20 \times 10^2$ \\
              & 1 & Raw          & 0.018 & $-928$& $3.01 \times 10^6$ & $2.65 \times 10^7$ \\
              & 1 & Clipped ($c=4$) & 5.72  & $-245$ & $1.15 \times 10^3$ & $1.04 \times 10^3$ \\
    \midrule
    Franka   & 0 & Raw          & 14.58 & 94.5  & 69.1                    & 0.495 \\
              & 1 & Raw          & 0.008 & $-819$& $1.91 \times 10^3$ & $1.09 \times 10^3$ \\
              & 1 & Clipped ($c=4$) & 9.46  & $-99.2$ & $1.87 \times 10^2$ & $1.49 \times 10^2$ \\
    \bottomrule
  \end{tabular}
\end{table*}

\begin{figure*}[h!]
  \centering
  \includegraphics[width=\textwidth]{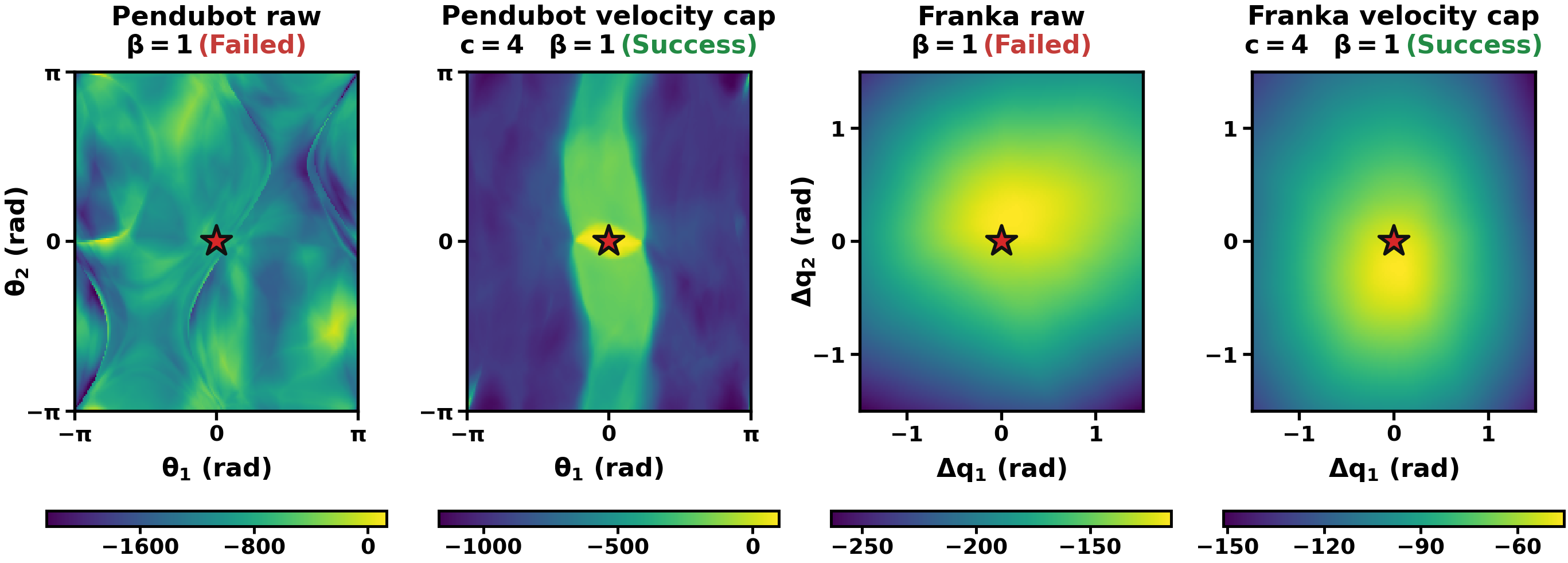}
  \caption{\textbf{Velocity capping contracts critic scale and restores
  target-centered structure.} From left to right: failed raw Pendubot,
  successful velocity-capped Pendubot, failed raw Franka, and successful
  velocity-capped Franka, all at $\beta=1$ and with $c=4$ in the capped
  conditions using seed 42. Pendubot fixes angular velocities at zero, while Franka holds all remaining
  joints at a common target-conditioned reference. Red stars mark the upright or zero-error target.}
  \label{fig:clipping_critics}
\end{figure*}

\Cref{fig:clipping_performance} shows that at $\beta=1$, \textbf{capping the velocity term substantially improves both Pendubot and Franka}. In \Cref{tab:clipping_probe}, the uncapped rewards produce much larger value MSE; the lower mean absolute minimum return target further shows that clipping reduces the leverage of rare high-speed samples in squared critic regression.
\Cref{fig:clipping_critics} shows the corresponding learned value functions. The raw Pendubot critic contains severe narrow extrapolations and a fragmented angle landscape, whereas the capped critic resolves a smooth upright-centered structure at an ordinary value scale. Franka's raw surface, associated with a failed controller, is smoother but has a broader negative scale; the capped critic concentrates its highest values around the target and accompanies a successful policy.

\paragraph{Full-pose constraints in high-DoF control.}

While the preceding results show that velocity penalties are not necessary for stabilization problems, a natural question is whether the reward needs to include all zeroth-order coordinates related to the goal. We examine this issue on the seven-DoF Franka arm by comparing two target specifications. Both retain the complete joint-position and joint-velocity observation and use the same position term and velocity sweep, \textbf{but they differ in which zeroth-order coordinates define the target}:
\begin{align}
    r^{\mathrm{Full}}_\beta
      &=1-3\tanh(d_{p}/0.8)
        -\beta\lVert\dot q\rVert_2^2 - 3e_q,
    \qquad
    e_{q} =\sqrt{\operatorname{mean}_j(q_{j}-q_j^*)^2},
    \label{eq:completeness_full_reward}\\
    r^{\mathrm{Partial}}_\beta
      &=1-3\tanh(d_{p}/0.8)-3d_{R}/\pi
        -\beta\lVert\dot q\rVert_2^2.
    \label{eq:completeness_partial_reward}
\end{align}
Here $d_p$ is end-effector position error and $d_R\in[0,\pi]$ is the geodesic end-effector orientation error (geodesic-angle form). The Full target joint posture $q^*$ is generated consistently with the desired end-effector pose. Partial explicitly constrains the complete end-effector pose (position and orientation) but not the remaining null-space coordinate. Thus, ``Partial" is complete in task space but incomplete in the robot's generalized configuration.

\begin{figure*}[h]
    \centering
    \includegraphics[width=\textwidth]{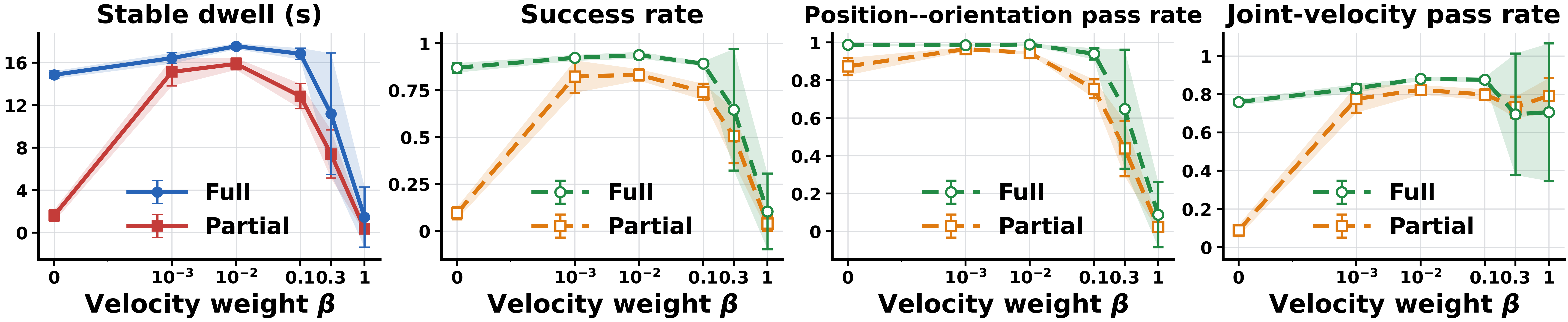}
    \caption{\textbf{Configuration completeness determines whether a zeroth-order reward is sufficient.} From left to right: best stable dwell (s), success-ever rate, position--orientation pass rate $P_{pR}$, and joint-velocity pass rate $P_v$. Full uses a target joint posture, whereas Partial specifies only the end-effector position and orientation. Curves and bands show the mean $\pm$ one population standard deviation over five seeds.}
    \label{fig:completeness}
\end{figure*}

For this ablation, let $C_p,C_R,C_v$ be the position, orientation, and joint-speed pass indicators specified in \Cref{tab:success_thresholds}. \Cref{eq:evaluation_metrics} applies with $C=C_pC_RC_v$; the two additional diagnostics are defined as time-step occupancies $P_{pR}=\mathbb E_{i,t}[C_pC_R]$ and $P_v=\mathbb E_{i,t}[C_v]$. Appendix~\ref{app:implement} gives more details.

Under the Partial setting at $\beta=0$, \textbf{performance collapses relative to Full} despite using the same complete observation. The decomposition in \Cref{fig:completeness} localizes the difference: Partial still occupies the desired end-effector pose, but rarely meets the joint-speed condition ($P_v=0.088\pm0.030$, versus $0.759\pm0.016$). Its principal failure is therefore continued null-space motion after reaching, not Cartesian reachability.
Under the Full setting, that motion changes $e_q$, so future zeroth-order rewards distinguish braking from joint drift; under the Partial setting, pose-preserving motion is reward-invariant and an explicit velocity term supplies the missing distinction. 

Accordingly, \textbf{a small} $\boldsymbol{\beta}\mathbf{=10^{-3}}$ \textbf{repairs Partial} to $15.13\pm1.31$ s dwell, and $P_v=0.775\pm0.071$. This comparison qualifies zeroth-order sufficiency: velocity reward is redundant for a configuration-complete target but can regularize directions that an incomplete target leaves invisible. As $\beta$ becomes dominant, both variants eventually degrade, so velocity regularization cannot replace configuration completeness and retains the large-weight optimization cost identified above.

\section{Conclusion and Future Works}

Our work provides a principled framework for reward design in deep reinforcement learning for robotics. We show that effective control does not require complex, multi-component reward functions burdened by higher-order penalties; rather, it relies on a clear separation between what the agent optimizes (zeroth-order target completeness) and what it observes (full state information). Omitting unnecessary velocity penalties eliminates extreme value tails and optimization brittleness, resulting in simpler, more robust policies without sacrificing performance. Moreover, these insights shift the paradigm of robotic reward shaping from trial-and-error heuristic tuning toward minimal, mathematically grounded design principles, paving the way for scalable control across increasingly complex and high-dimensional physical systems.

It should be noted that our study is limited to simulated stabilization and target-reaching tasks in second-order systems, trained with feedforward PPO. Moreover, passive damping, low-level feedback, and policy memory can change which information must be supplied explicitly.
These limitations motivate an important open problem: whether this separation extends to higher-order and higher-dimensional systems. For a state containing $q,q^{(1)},\ldots,q^{(d-1)}$, what are the minimal reward and observation orders needed to evaluate and realize the desired trajectories? Establishing mode-wise conditions and adaptive shaping that adds only unresolved derivative information may be promising directions for future work.

\section*{Reproducibility Statement}

We provide the information needed to reproduce every result. Theorem~\ref{th:prop1} and Assumption~\ref{assumption} appear in Section~\ref{sec:method}, with the trajectory-level derivation, numerical-solver analysis, and proof in Appendices~\ref{app:hessian}, \ref{app:trajectory_level_view}, and~\ref{app:proof}, respectively. The task rewards and all success predicates are in \Cref{tab:isaaclab_reward_family} and \Cref{tab:success_thresholds}; the reward-clipping and reward-completeness interventions are given as explicit formulas in \Cref{eq:main_velocity_cap}--\Cref{eq:franka_velocity_clip} and \Cref{eq:completeness_full_reward}--\Cref{eq:completeness_partial_reward}. PPO hyperparameters, architectures, and budgets are listed in Appendix~\ref{app:ppo_hyperparameters} (\Cref{tab:ppo_runner_network,tab:ppo_algorithm}). We report Isaac Lab aggregate curves over five seeds $\{0,1,2,3,42\}$.

\section*{AI Use Statement}

Generative AI tools, including ChatGPT 5.6 Sol (OpenAI) and Claude Opus 5 (Anthropic), were used to improve the clarity, grammar, and presentation of author-written text, and to assist with literature retrieval and discovery by suggesting search queries and potentially relevant prior work. All cited references were independently located and verified against their original publications by the authors. All AI-assisted edits were reviewed and revised by the authors. The authors take full responsibility for the final content of this work, including any text produced with the aid of generative AI.

\bibliography{iclr2027_conference}
\bibliographystyle{iclr2027_conference}

\newpage
\appendix

\section{Implementation Details}
\label{app:implement}

\paragraph{Environments.}

\begin{figure*}[h!]
    \centering
    \begin{subfigure}[b]{0.24\textwidth}
        \centering
        \includegraphics[width=\textwidth]{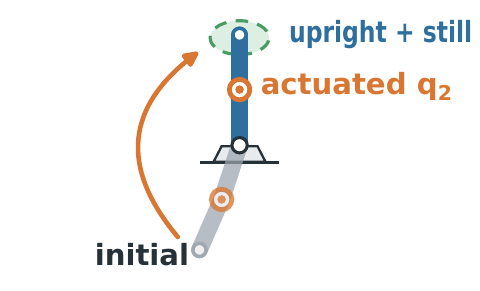}
        \caption{Acrobot}
        \label{fig:task_a}
    \end{subfigure}
    \hfill
    \begin{subfigure}[b]{0.24\textwidth}
        \centering
        \includegraphics[width=\textwidth]{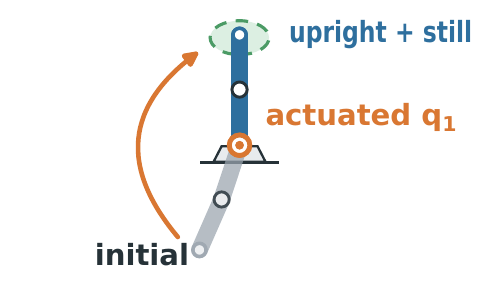}
        \caption{Pendubot}
        \label{fig:task_p}
    \end{subfigure}
    \hfill
    \begin{subfigure}[b]{0.24\textwidth}
        \centering
        \includegraphics[width=\textwidth]{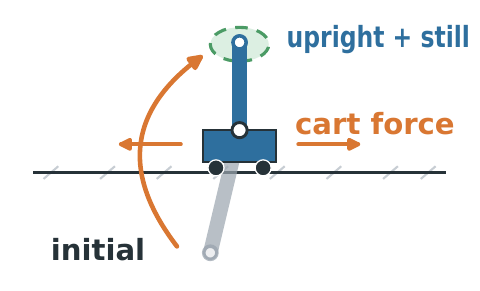}
        \caption{Cartpole}
        \label{fig:task_c}
    \end{subfigure}
    \hfill
    \begin{subfigure}[b]{0.24\textwidth}
        \centering
        \includegraphics[width=\textwidth]{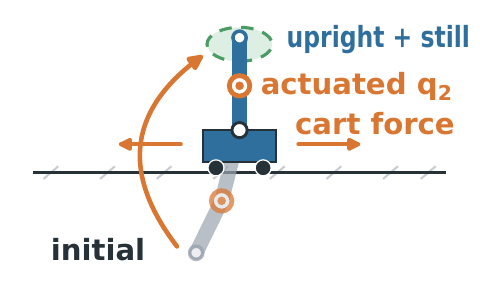}
        \caption{Double Cart Pendulum}
        \label{fig:task_d}
    \end{subfigure}
    
    \begin{subfigure}[b]{0.24\textwidth}
        \centering
        \includegraphics[width=\textwidth]{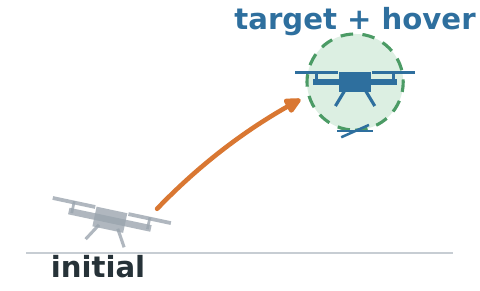}
        \caption{Quadcopter}
        \label{fig:task_q}
    \end{subfigure}
    \begin{subfigure}[b]{0.24\textwidth}
        \centering
        \includegraphics[width=\textwidth]{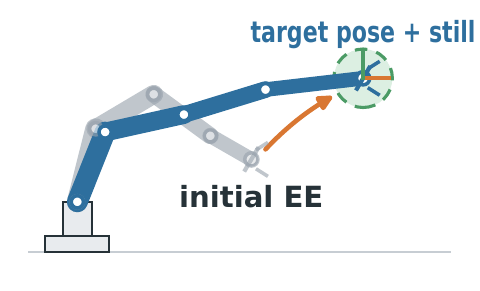}
        \caption{Franka Reach}
        \label{fig:task_f}
    \end{subfigure}
    \begin{subfigure}[b]{0.24\textwidth}
        \centering
        \includegraphics[width=\textwidth]{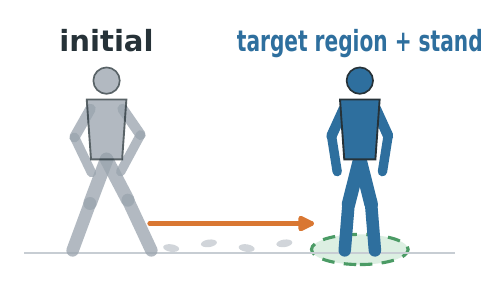}
        \caption{Humanoid}
        \label{fig:task_h}
    \end{subfigure}

    \caption{Environments.}
    \label{fig:isaaclab_tasks}
\end{figure*}

\begin{figure*}[h!]
    \centering
    \begin{subfigure}[b]{0.24\textwidth}
        \centering
        \includegraphics[width=\textwidth]{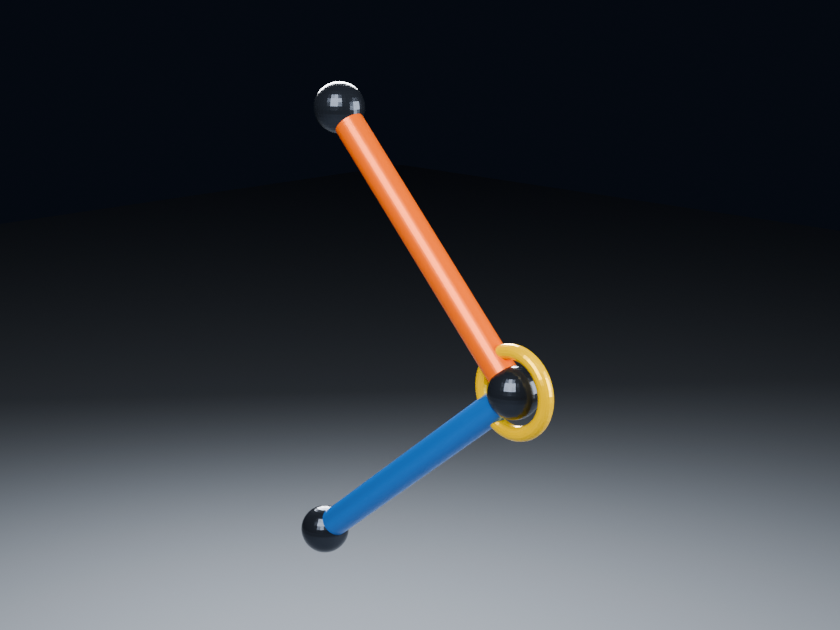}
        \caption{Acrobot}
        \label{fig:rendered_task_a}
    \end{subfigure}
    \hfill
    \begin{subfigure}[b]{0.24\textwidth}
        \centering
        \includegraphics[width=\textwidth]{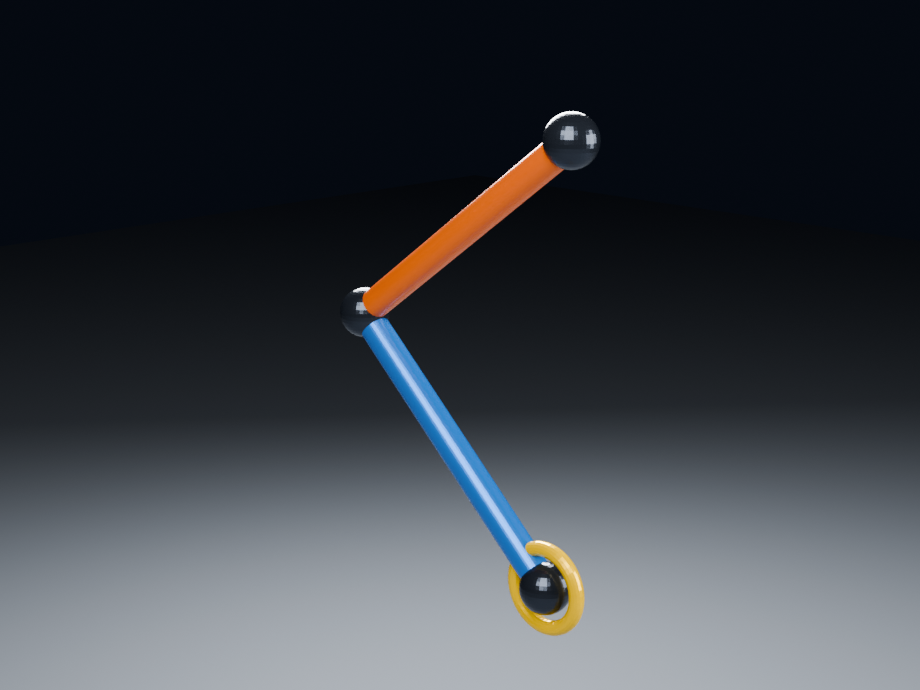}
        \caption{Pendubot}
        \label{fig:rendered_task_p}
    \end{subfigure}
    \hfill
    \begin{subfigure}[b]{0.24\textwidth}
        \centering
        \includegraphics[width=\textwidth]{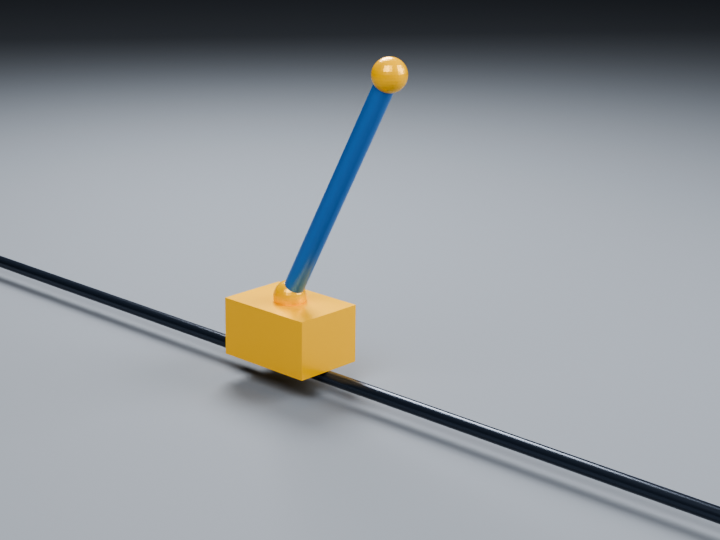}
        \caption{Cartpole}
        \label{fig:rendered_task_c}
    \end{subfigure}
    \hfill
    \begin{subfigure}[b]{0.24\textwidth}
        \centering
        \includegraphics[width=\textwidth]{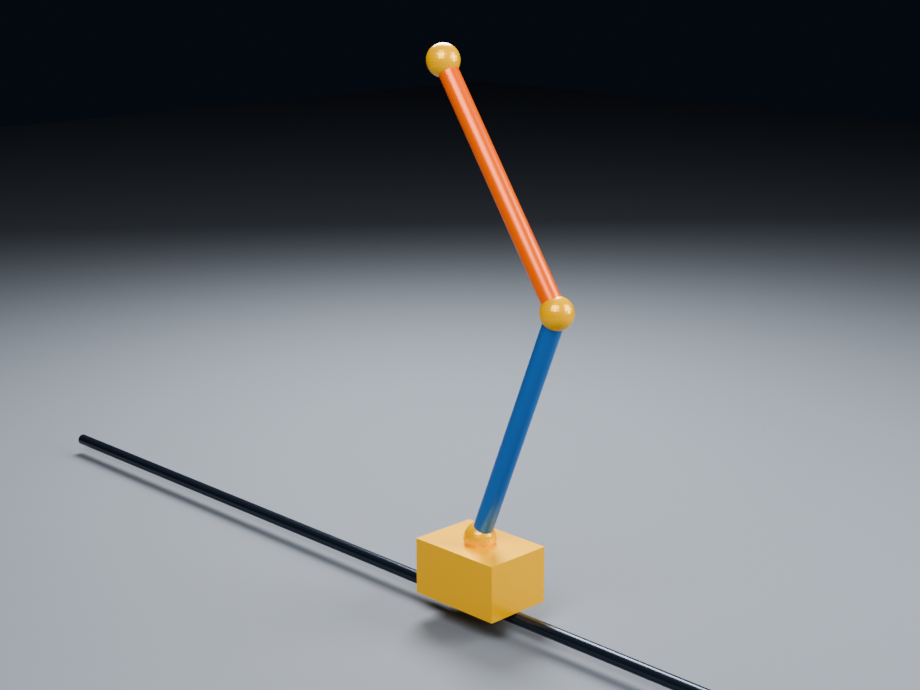}
        \caption{Double Cart Pendulum}
        \label{fig:rendered_task_d}
    \end{subfigure}

    \begin{subfigure}[b]{0.24\textwidth}
        \centering
        \includegraphics[width=\textwidth]{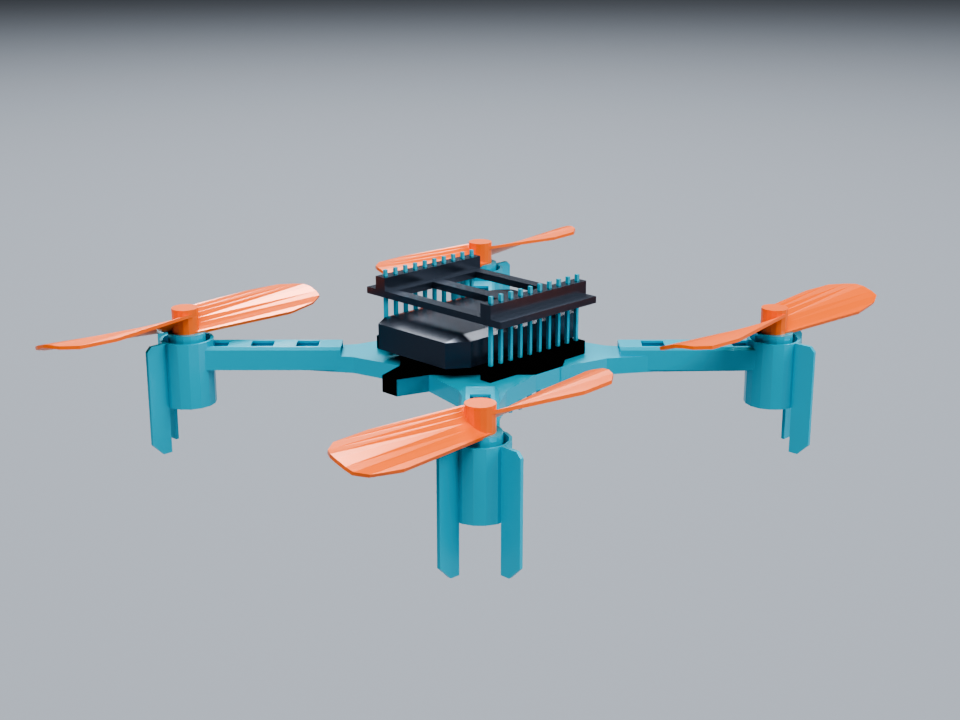}
        \caption{Quadcopter}
        \label{fig:rendered_task_q}
    \end{subfigure}
    \begin{subfigure}[b]{0.24\textwidth}
        \centering
        \includegraphics[width=\textwidth]{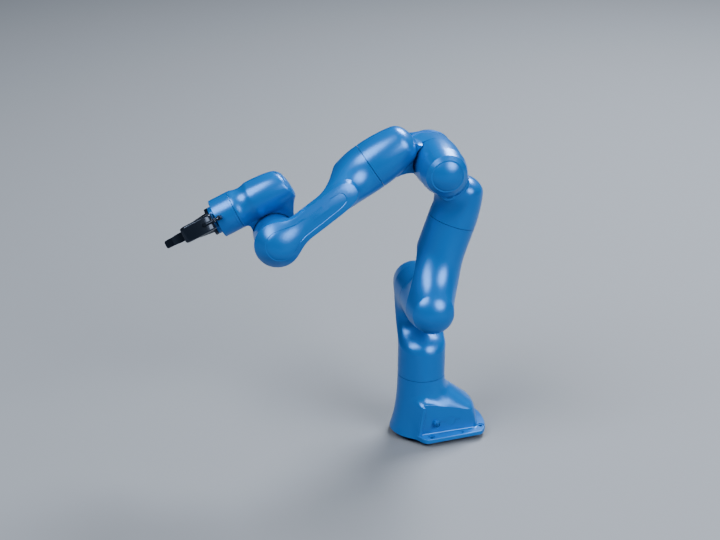}
        \caption{Franka Reach}
        \label{fig:rendered_task_f}
    \end{subfigure}
    \begin{subfigure}[b]{0.24\textwidth}
        \centering
        \includegraphics[width=\textwidth]{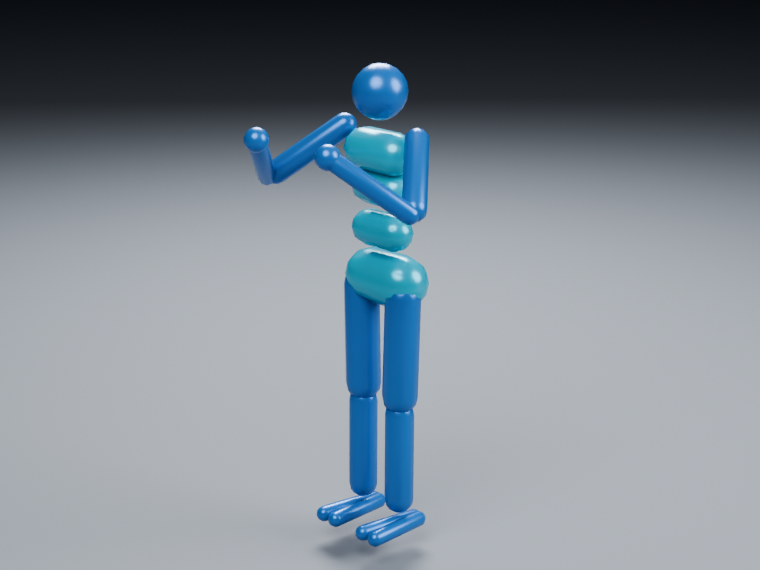}
        \caption{Humanoid}
        \label{fig:rendered_task_h}
    \end{subfigure}

    \caption{Rendered views of the seven environments.}
    \label{fig:isaaclab_tasks_rendered}
\end{figure*}

The seven Isaac Lab tasks used are shown in Figures~\ref{fig:isaaclab_tasks} and~\ref{fig:isaaclab_tasks_rendered}. The suite ranges from low-dimensional underactuated mechanisms to floating-base, manipulation, and contact-rich whole-body systems.

\paragraph{Deterministic trajectory evaluation.}

We evaluate the deterministic policy mean for 600 control steps using a fixed bank of $N=128$ initial conditions or targets ($N=640$ for Humanoid). The common metrics are defined in \Cref{eq:evaluation_metrics}; \Cref{tab:success_thresholds} gives every instantaneous predicate used to construct $C_{i,t}$.

\begin{table*}[h]
\centering
\caption{Instantaneous success predicates for all Isaac Lab evaluations. All
conditions in a row must hold simultaneously.}
\label{tab:success_thresholds}
\small
\setlength{\tabcolsep}{5pt}
\begin{tabular}{@{}lp{0.7\textwidth}@{}}
\toprule
Task & Conditions for $C_{i,t}=1$ \\
\midrule
Acrobot / Pendubot &
$|\theta_1|, |\theta_{2,a}|<0.15$ rad, $|\dot\theta_1|, |\dot\theta_2|<0.5$ rad/s. \\
Cartpole &
$|\theta|<0.15$ rad, $|v_c|<0.5$ m/s, $|\dot\theta|<0.5$ rad/s. \\
Double Cart Pendulum &
$|\theta_1|, |\theta_{2,a}|<0.15$ rad, $|v_c|<0.5$ m/s, $|\dot\theta_1|, |\dot\theta_2|<0.5$ rad/s\\
Quadcopter &
$d_p<0.10$ m, $\lVert v_{\mathrm{body}}\rVert_2<0.20$ m/s, $\lVert\omega_{\mathrm{body}}\rVert_2<0.30$ rad/s, tilt angle $<0.15$ rad. \\
Franka Reach (Full / Partial) &
$d_p<0.05$ m, $d_R<0.20$ rad, $\lVert\dot q\rVert_2<0.25$ rad/s. \\
Humanoid &
$d_{xy}<0.35$ m, $\lVert v_{\mathrm{root}}\rVert_2<0.35$ m/s, $\lVert\omega_{\mathrm{root}}\rVert_2<0.75$ rad/s. \\
\bottomrule
\end{tabular}
\end{table*}

\paragraph{Reward clipping.}

We use reward clipping in the targeted form shown in \Cref{eq:main_velocity_cap}: it clips the velocity error rather than the whole reward. In the notation of \Cref{tab:isaaclab_reward_family}, both tasks use
\begin{equation}
    r_{\beta,c}(q,v)=r_0(q)-\beta e_v^{(c)}(v),
    \qquad e_v^{(c)}(v)=\min\{e_v(v),c\},
    \qquad c=4.
    \label{eq:generic_velocity_cap}
\end{equation}
In particular, the zeroth-order term and the total reward remain unclipped, so the instantaneous velocity penalty is at most $\beta c$. It preserves the unique minimum of the velocity error at $e_v=0$ while removing its unbounded contribution to the reward's left tail under stochastic exploration. At $\beta=0$ it is algebraically inactive and the capped and raw reward functions are identical.

For Pendubot, let $\theta_{2,a}=\theta_1+\theta_2$ be the absolute second-link angle and define $e_c(\theta_1)=1-\cos\theta_1$, $e_c(\theta_{2,a})=1-\cos\theta_{2,a}$. The two rewards are
\begin{align}
    r_{\beta}^{\mathrm P}
      &=1-3[e_c(\theta_1) + e_c (\theta_{2,a})]
        -\beta\bigl(\dot\theta_1^2+\dot\theta_2^2\bigr),
        \label{eq:pendubot_raw_reward}\\
    r_{\beta,c}^{\mathrm P}
      &=1-3[e_c(\theta_1) + e_c (\theta_{2,a})]
        -\beta\min \left(\dot\theta_1^2+\dot\theta_2^2,c\right),
        \qquad c=4.
        \label{eq:pendubot_velocity_cap}
\end{align}

For Franka, let $d_p$ be the end-effector position error and $e_q=\sqrt{\operatorname{mean}_j(q_j-q_j^*)^2}$ be the arm joint-posture error. We analogously compare
\begin{align}
    r_{\beta}^{\mathrm F}
      &=1-3\tanh(d_p/0.8)-3e_q-\beta\lVert\dot q\rVert_2^2,
      \label{eq:franka_raw_reward}\\
    r_{\beta,c}^{\mathrm F}
      &=1-3\tanh(d_p/0.8)-3e_q
        -\beta\min \left(\lVert\dot q\rVert_2^2,c\right),
        \qquad c=4.
      \label{eq:franka_velocity_clip}
\end{align}

All other reward, observation, PPO, training, and evaluation settings are held fixed between the raw and reward-clipped sweeps. We use five seeds $\{0,1,2,3,42\}$ for this experiment; the seed-42 one-iteration diagnostics in \Cref{tab:clipping_probe} average saved checkpoints $\{1800,1950,2100,2250,2299\}$.

\paragraph{Reward order versus reward completeness.}

Now we formalize the detailed theory of the necessity of reward completeness. Let $y=h(q)$ be the zeroth-order task output and rewrite the zeroth-order reward term as $r_0 = r_0(h(q))$. If $h$ is non-injective, then distinct configurations can receive exactly the same instantaneous reward.
For a redundant manipulator with Jacobian $J_h(q)$, any local displacement $\delta q\in\ker J_h(q)$ satisfies
\begin{equation}
    h(q+\delta q)=h(q)+O(\lVert\delta q\rVert_2^2),
    \qquad
    \nabla_q r_0(h(q))^\top\delta q=0.
    \label{eq:reward_nullspace}
\end{equation}
Finite self-motion manifolds can make this ambiguity exact. A trajectory return composed only of $r_0(h(q))$ cannot distinguish two trajectories that share the same reward value while moving differently within such a local set, making it impossible to rank behaviors that the selected $r_0$ leaves indistinguishable.

The Full and Partial rewards and their empirical comparison are given in \Cref{eq:completeness_full_reward}--\Cref{eq:completeness_partial_reward} and \Cref{fig:completeness}; all observations, PPO settings, and evaluation trajectories are otherwise matched.

Under the Partial setting, the task-map Jacobian generically leaves at least one tangent direction for a seven-DoF arm away from singularities. Motion along the associated self-motion manifold can preserve the full end-effector pose over time, so not only the instantaneous reward but also future pose rewards can remain unchanged.
The Full posture term removes this ambiguity locally by making any generalized-coordinate departure from $q^*$ costly. A positive velocity penalty does not make the Partial target injective; it instead ranks motion along the otherwise reward-equivalent manifold and thereby supplies direct damping. This geometric distinction explains why a small velocity weight can compensate for Partial's missing configuration constraint without making velocity reward intrinsically necessary for the Full target.

\section{PPO Hyperparameters}
\label{app:ppo_hyperparameters}

\Cref{tab:ppo_runner_network,tab:ppo_algorithm} reports the effective configuration used for the beta sweeps and no-velocity-observation ablations. The first column applies to Acrobot, Pendubot, Cartpole, Double Cart Pendulum, Quadcopter, and Franka; Humanoid-specific values are shown separately. Unless listed otherwise, the two groups use the same setting.

\begin{table*}[h]
\centering
\caption{Rollout, runner, and actor--critic settings.}
\label{tab:ppo_runner_network}
\small
\begin{tabular}{@{}lll@{}}
\toprule
Parameter & Six-task shared setting & Humanoid setting \\
\midrule
Parallel environments & 4096 & 4096 \\
Rollout steps per environment & 600 & 600 \\
Transitions per PPO update & 2,457,600 & 2,457,600 \\
Training iterations & 2300 & 1500 \\
Checkpoint interval / final checkpoint & 150 / 2299 & 150 / 1499 \\
Training seeds & $0,1,2,3,42$ & $0,1,2,3,42$ \\
Observation normalization & empirical, enabled & empirical, enabled \\
Policy class & feedforward Gaussian actor--critic & same \\
Actor hidden widths & $256,256,256$ & $400,200,100$ \\
Critic hidden widths & $256,256,256$ & $400,200,100$ \\
Activation & Leaky-ReLU & ELU \\
Critic output & one scalar value head & one scalar value head \\
Initial policy standard deviation (learned thereafter) & 0.7 & 1.0 \\
Standard-deviation parameterization & scalar & scalar \\
Standard-deviation floor & disabled & disabled \\
Runner action clipping & not set & 1.0 \\
\bottomrule
\end{tabular}
\end{table*}

\begin{table*}[h]
\centering
\caption{PPO objective and optimization settings.}
\label{tab:ppo_algorithm}
\small
\begin{tabular}{@{}lll@{}}
\toprule
Parameter & Six-task shared setting & Humanoid setting \\
\midrule
Optimizer & Adam & Adam \\
Initial learning rate & $10^{-3}$ & $10^{-4}$ \\
Learning-rate schedule & adaptive KL & adaptive KL \\
Desired KL & 0.01 & 0.008 \\
Policy ratio clip $\epsilon$ & 0.2 & 0.2 \\
Learning epochs per rollout & 5 & 5 \\
Minibatches per epoch & 4 & 4 \\
Transitions per minibatch & 614,400 & 614,400 \\
Discount $\gamma$ & 0.99 & 0.99 \\
GAE $\lambda$ & 0.95 & 0.95 \\
Value-loss coefficient & 1.0 & 1.0 \\
Clipped value loss & enabled & enabled \\
Entropy coefficient & 0.005 & 0.0 \\
Maximum gradient norm & 1.0 & 1.0 \\
Symmetry augmentation / RND & none / none & none / none \\
\bottomrule
\end{tabular}
\end{table*}

\section{Additional Experiment Results}
\label{app:additional}

For training iteration $k$, let $\mathcal{B}_k$ be the set of collected episodes and let $T$ be the length of episodes. We measure the Mean Return as
\begin{equation}
    \overline{G}_k
    = \frac{1}{|\mathcal{B}_k|}
      \sum_{\tau\in\mathcal{B}_k}
      \sum_{t=0}^{T-1} r_\beta(q_t,v_t).
    \label{eq:appendix_mean_return}
\end{equation}
Additionally, for the critic, let $\widehat G_t$ denote the bootstrapped return target in a PPO update and $V_{\phi_{\mathrm{old}}}(s_t)$ the value stored when the rollouts were collected. The logged Value Loss is
\begin{equation}
    \mathcal{L}_{V,k}
    = \frac{1}{|\mathcal{B}_k|}
      \sum_{t\in\mathcal{B}_k}
      \max\left\{
        \left(V_\phi(s_t)-\widehat G_t\right)^2,
        \left(V_{\phi}^{\mathrm{clip}}(s_t)-\widehat G_t\right)^2
      \right\},
    \label{eq:appendix_value_loss}
\end{equation}
where $V_{\phi}^{\mathrm{clip}}(s_t) = V_{\phi_{\mathrm{old}}}(s_t) + \operatorname{clip}\left( V_\phi(s_t)-V_{\phi_{\mathrm{old}}}(s_t),-\eta,\eta \right)$.

For visualization, each seed trace is first processed with a centered 51-iteration moving average, after which we report the mean and standard deviation across seeds $\{0,1,2,3,42\}$.

\begin{figure*}[h]
    \centering
    \setlength{\trainingcurveheight}{0.194\textwidth}
    \begin{subfigure}[b]{0.259\textwidth}
        \centering
        \includegraphics[height=\trainingcurveheight]{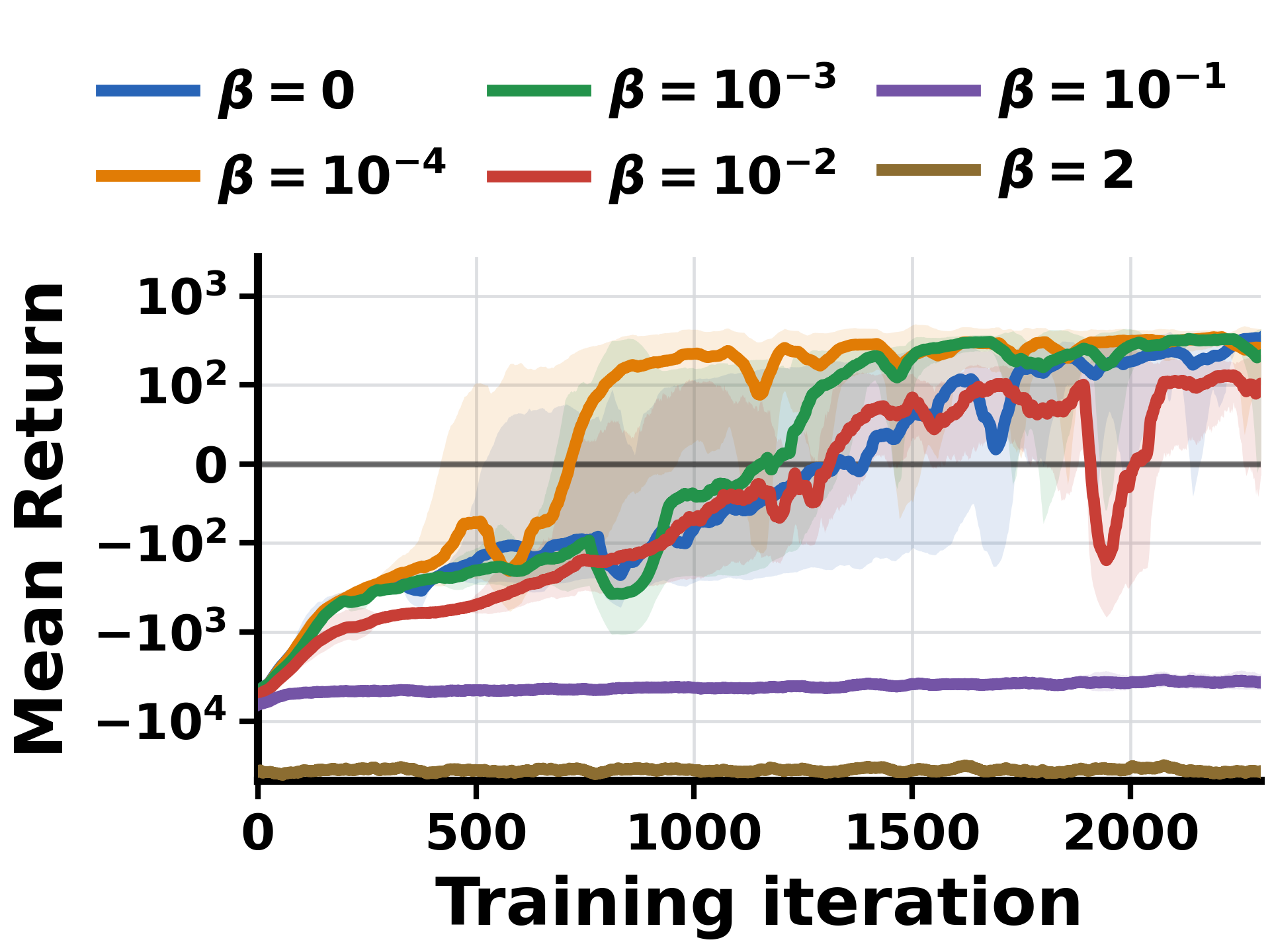}
        \caption{Acrobot}
        \label{fig:training_return_a}
    \end{subfigure}%
    \begin{subfigure}[b]{0.243\textwidth}
        \centering
        \includegraphics[height=\trainingcurveheight]{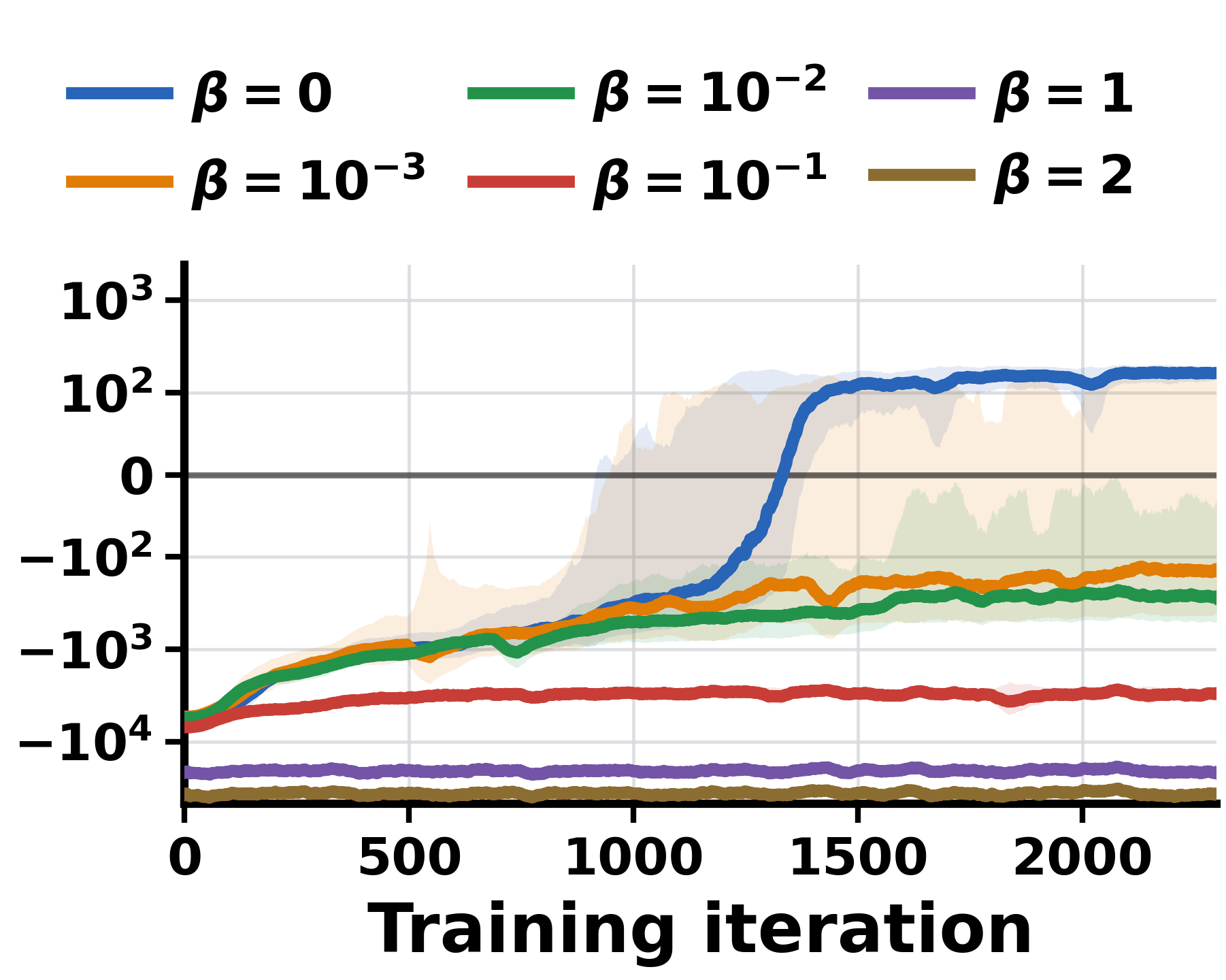}
        \caption{Pendubot}
        \label{fig:training_return_p}
    \end{subfigure}%
    \begin{subfigure}[b]{0.243\textwidth}
        \centering
        \includegraphics[height=\trainingcurveheight]{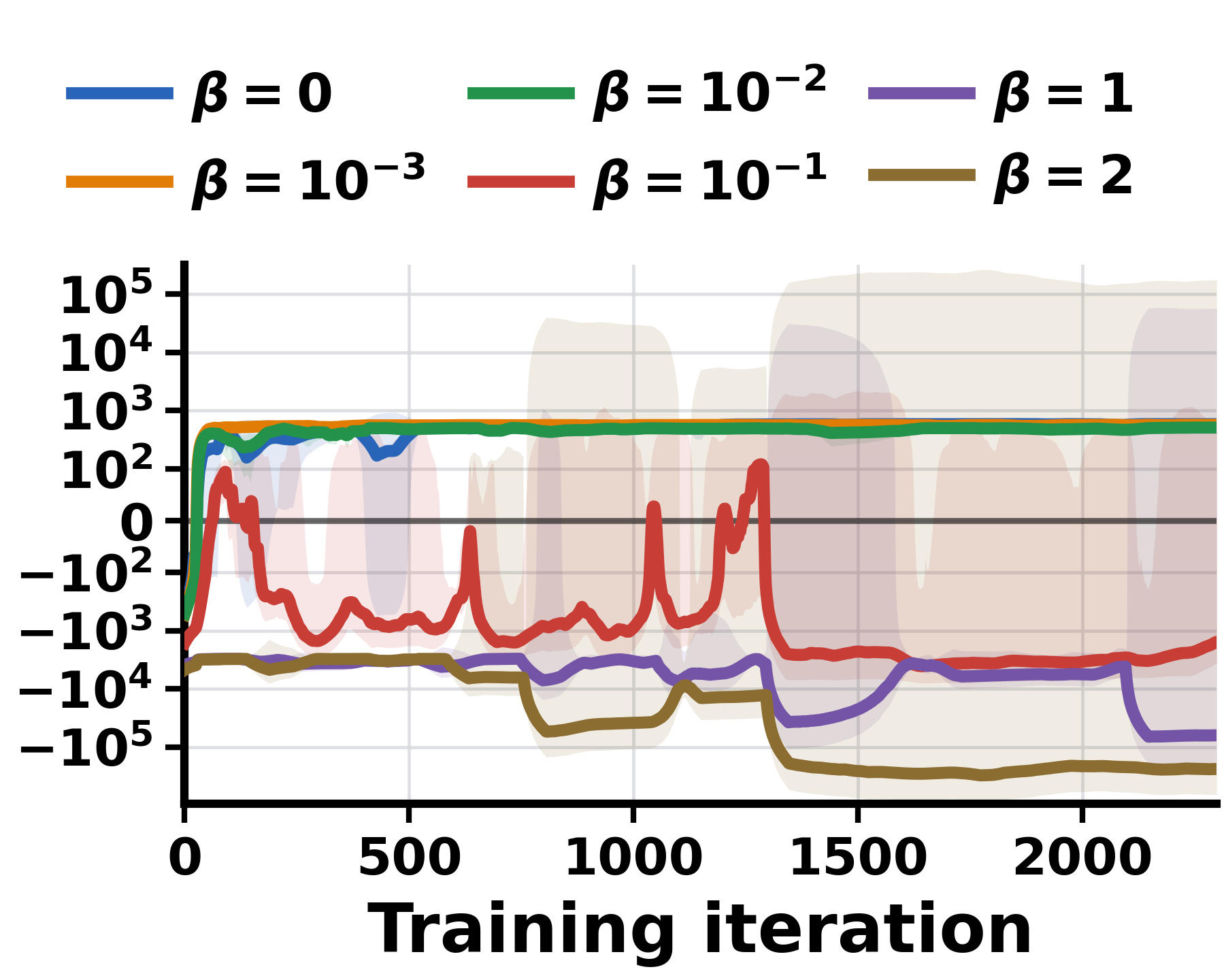}
        \caption{Cartpole}
        \label{fig:training_return_c}
    \end{subfigure}%
    \begin{subfigure}[b]{0.243\textwidth}
        \centering
        \includegraphics[height=\trainingcurveheight]{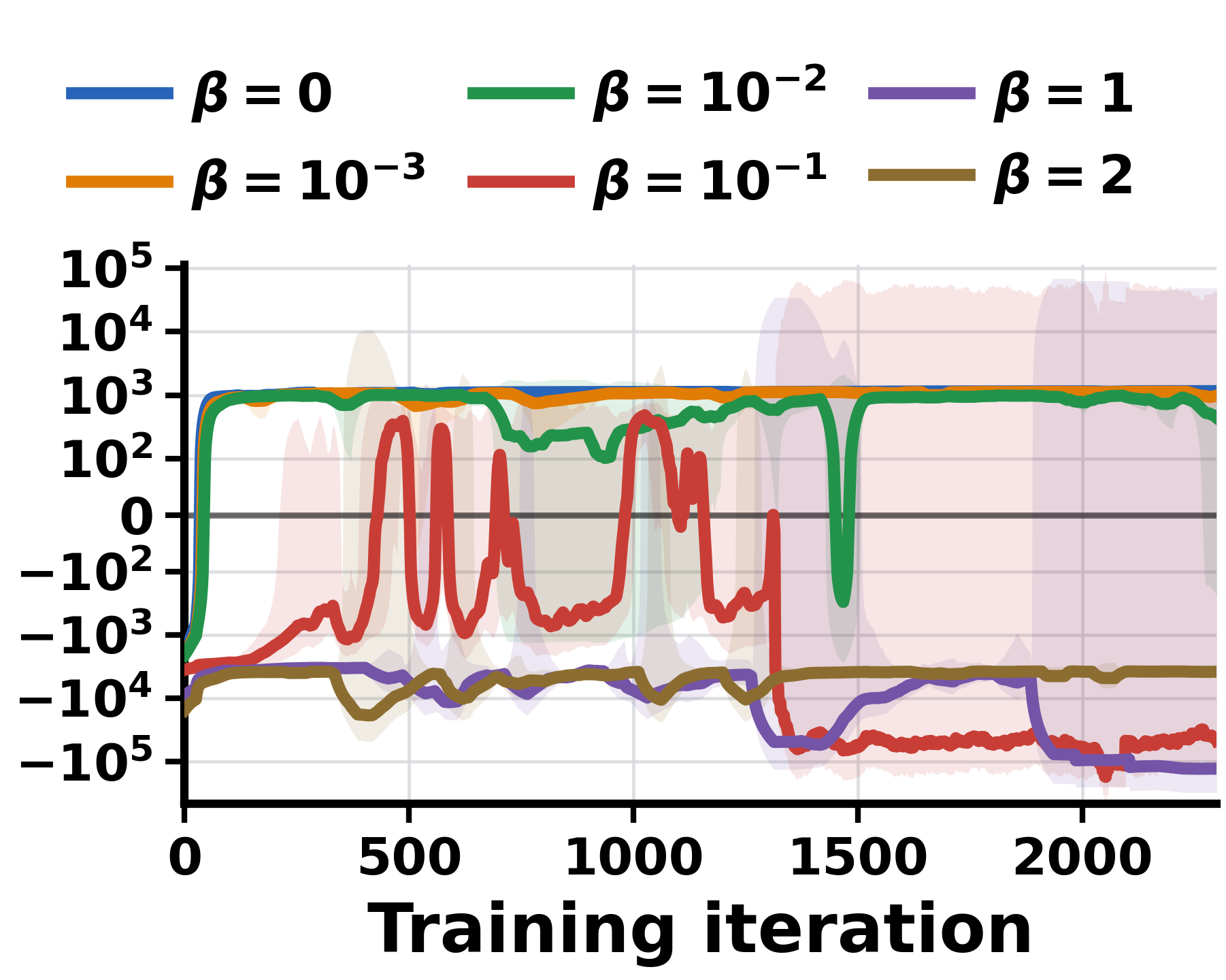}
        \caption{Double Cart Pendulum}
        \label{fig:training_return_d}
    \end{subfigure}

    \begin{subfigure}[b]{0.259\textwidth}
        \centering
        \includegraphics[height=\trainingcurveheight]{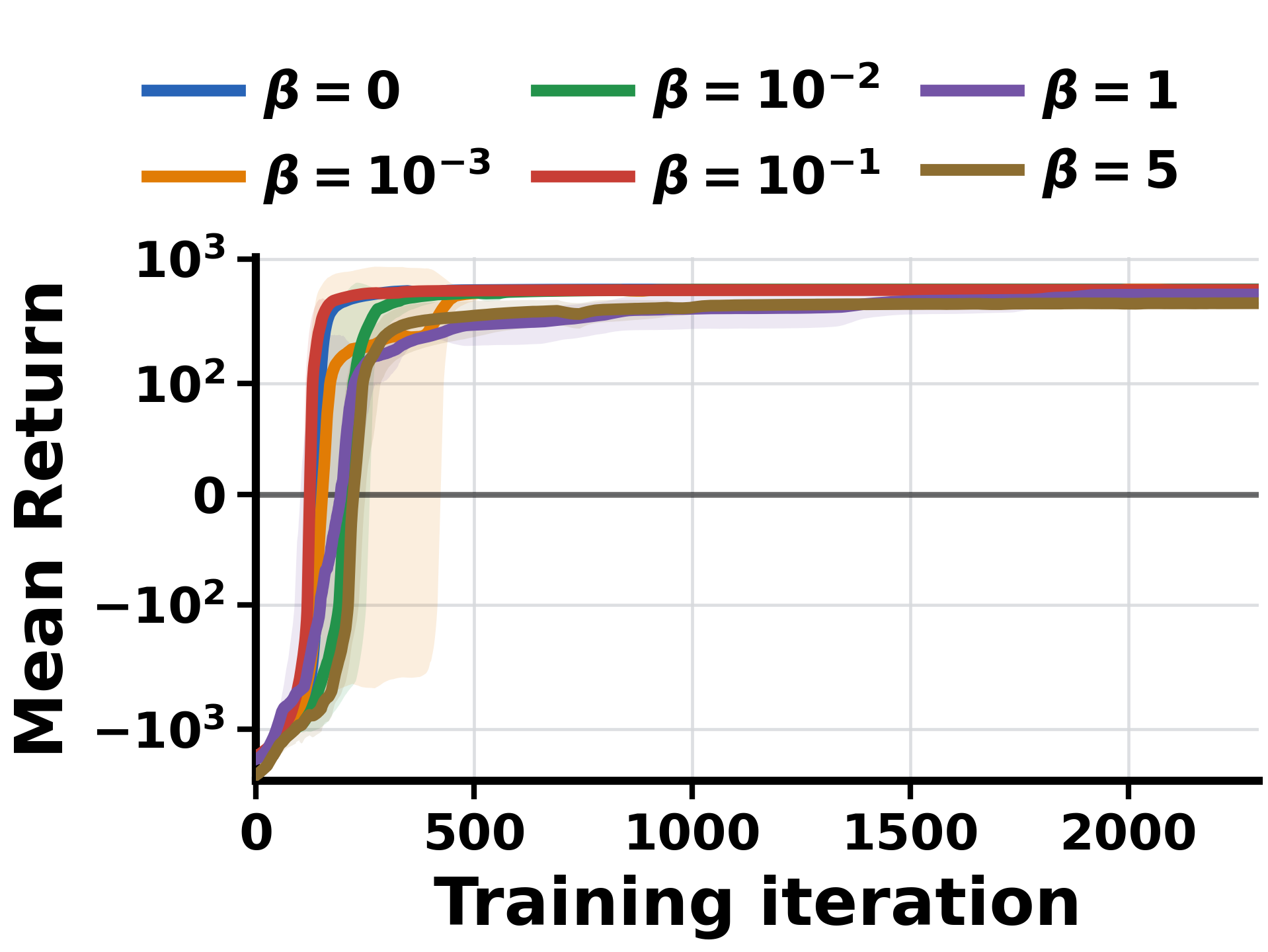}
        \caption{Quadcopter}
        \label{fig:training_return_q}
    \end{subfigure}%
    \begin{subfigure}[b]{0.243\textwidth}
        \centering
        \includegraphics[height=\trainingcurveheight]{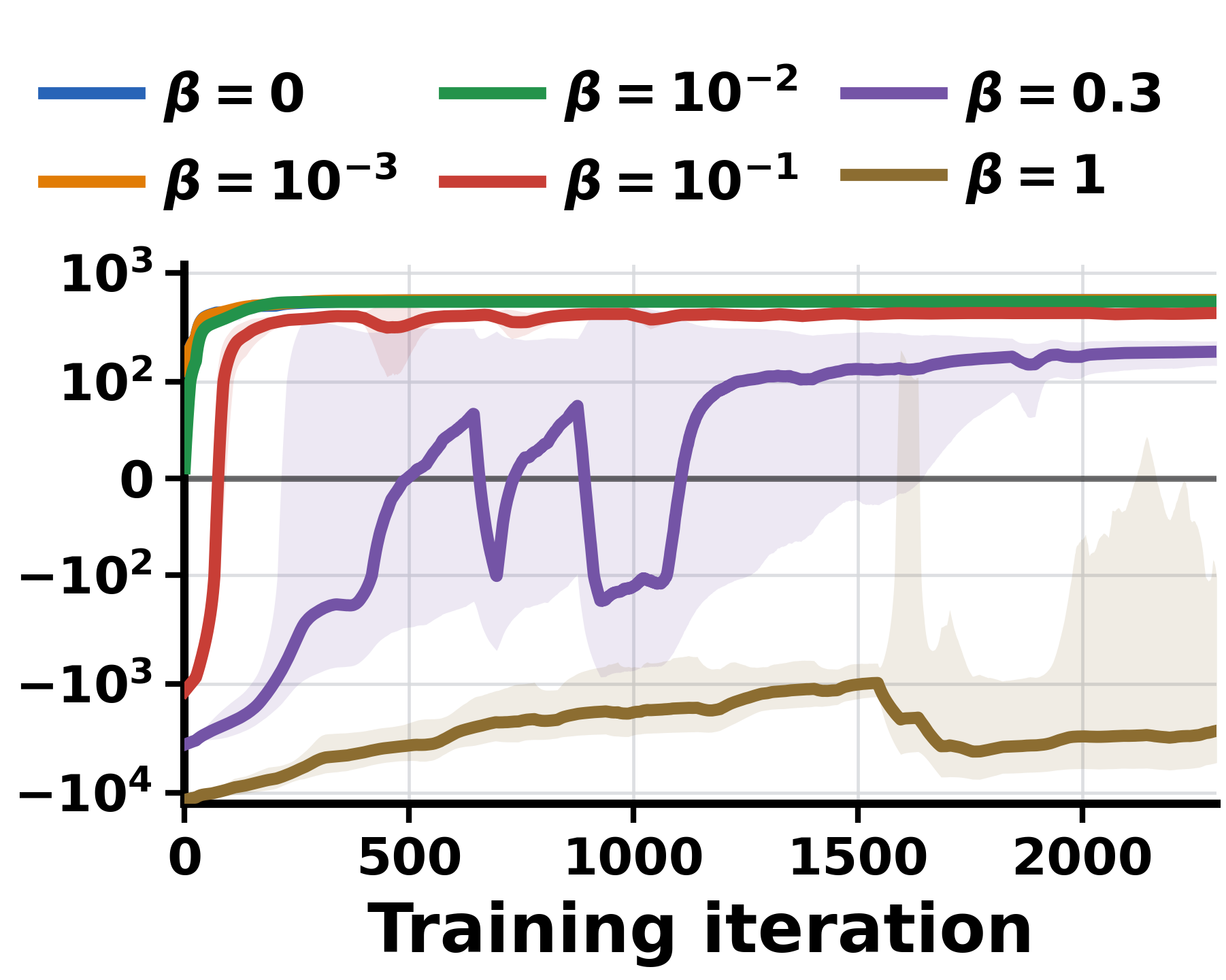}
        \caption{Franka Reach}
        \label{fig:training_return_f}
    \end{subfigure}%
    \begin{subfigure}[b]{0.243\textwidth}
        \centering
        \includegraphics[height=\trainingcurveheight]{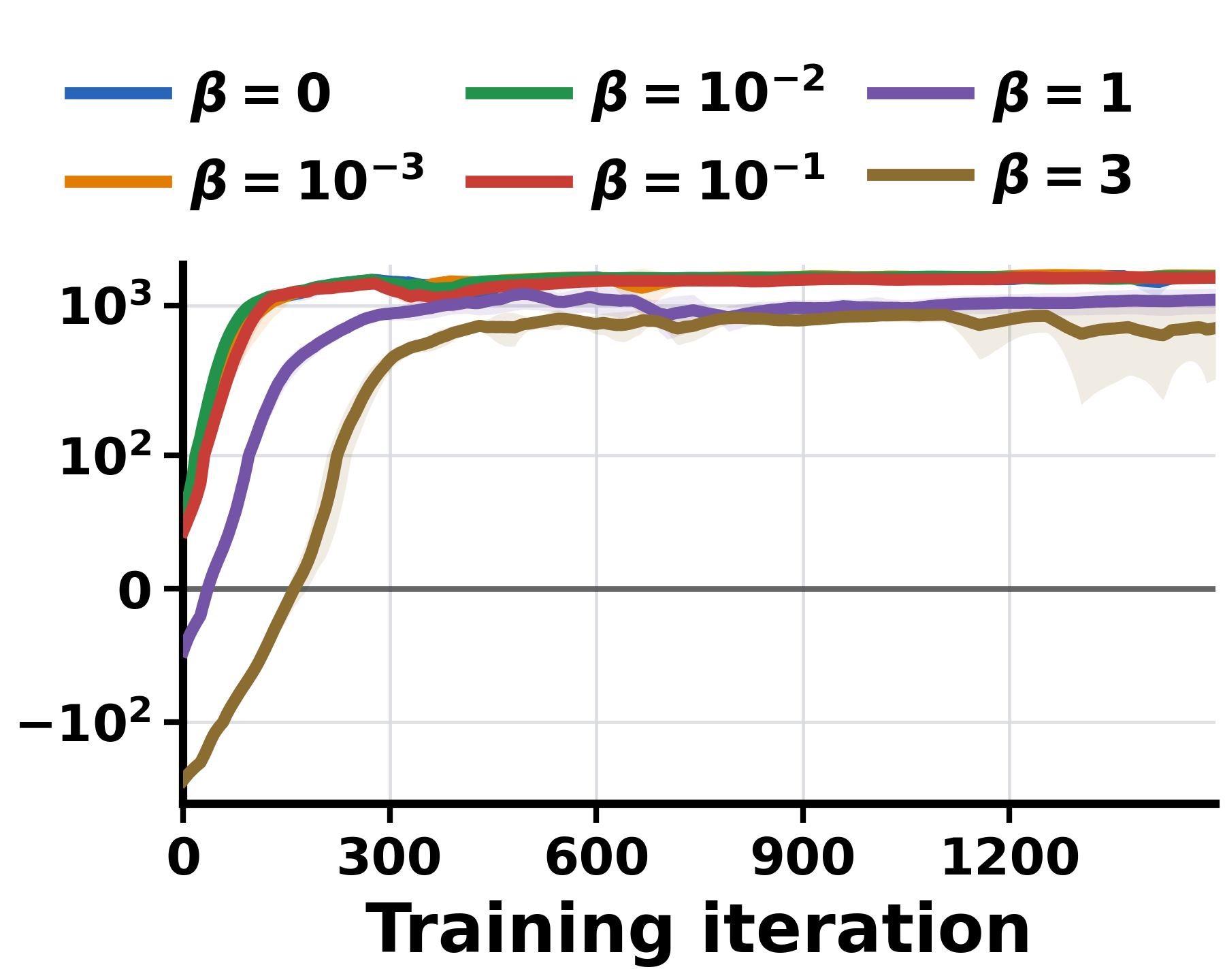}
        \caption{Humanoid}
        \label{fig:training_return_h}
    \end{subfigure}

    \caption{Training return (\Cref{eq:appendix_mean_return}) across the seven Isaac Lab tasks. Each seed is first smoothed with a centered 51-iteration moving average; curves and shaded regions then show the across-seed mean and one standard deviation, respectively.}
    \label{fig:training_curves_return}
\end{figure*}

\begin{figure*}[h]
    \centering
    \setlength{\trainingcurveheight}{0.198\textwidth}
    \begin{subfigure}[b]{0.256\textwidth}
        \centering
        \includegraphics[height=\trainingcurveheight]{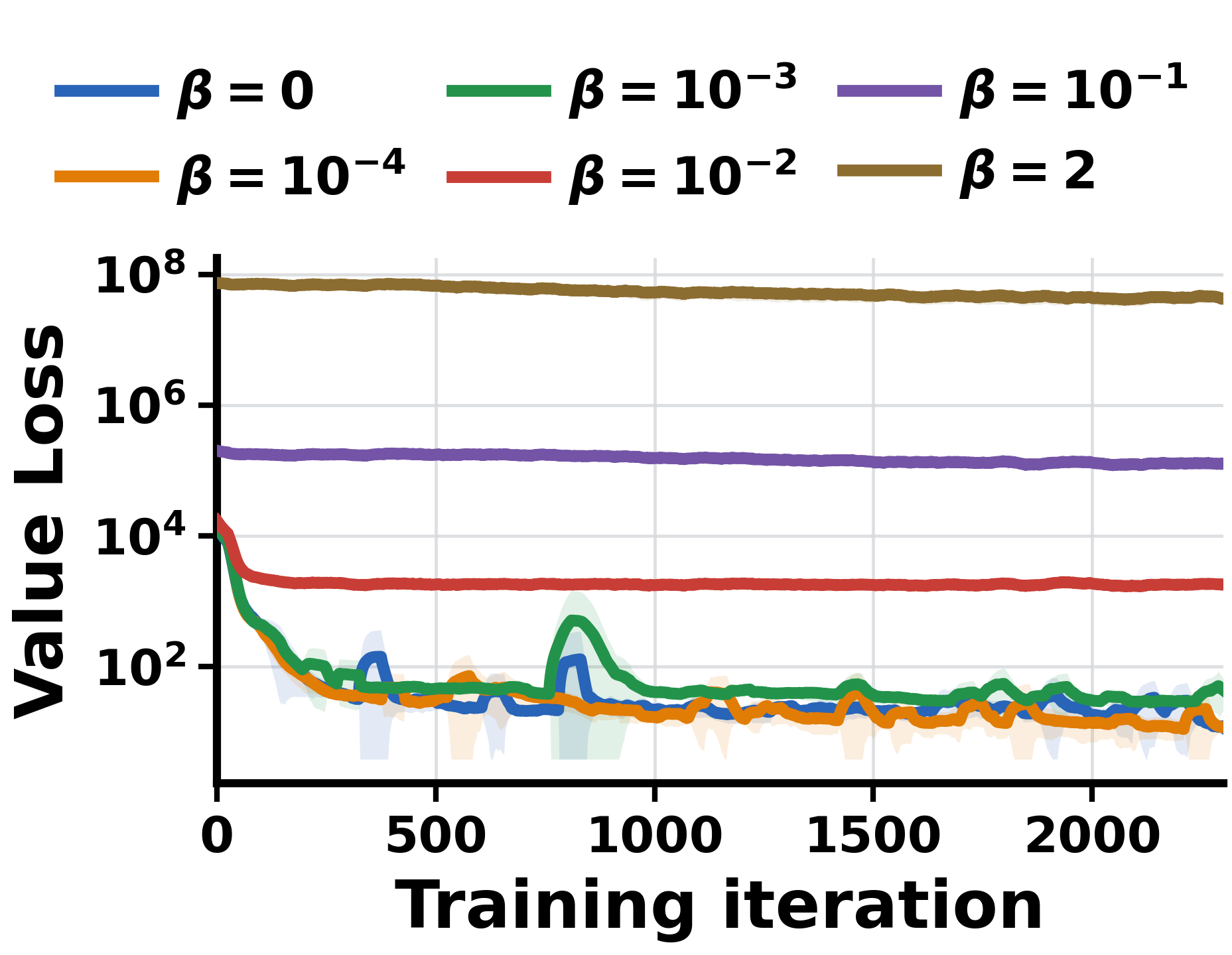}
        \caption{Acrobot}
        \label{fig:training_loss_a}
    \end{subfigure}%
    \begin{subfigure}[b]{0.241\textwidth}
        \centering
        \includegraphics[height=\trainingcurveheight]{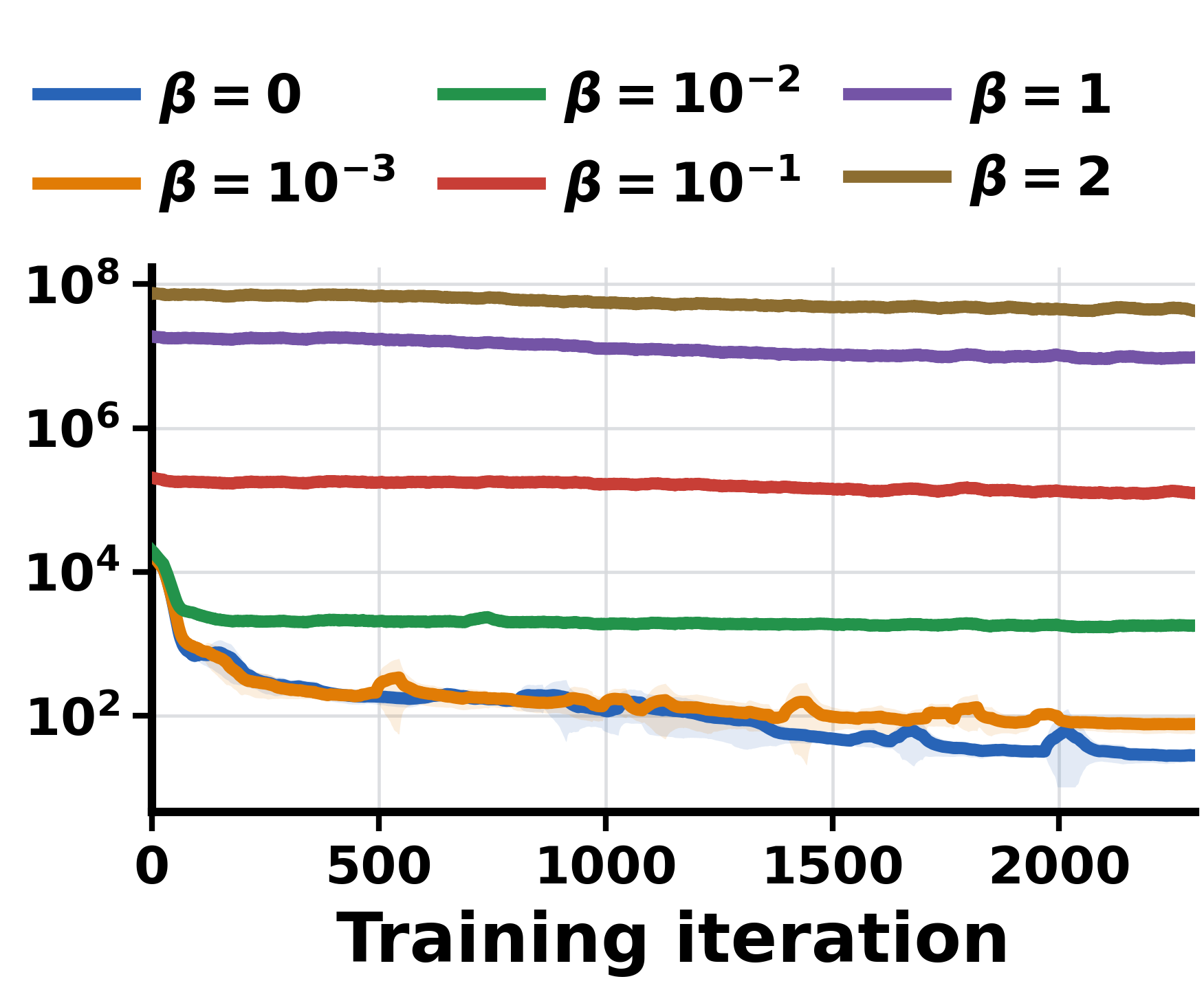}
        \caption{Pendubot}
        \label{fig:training_loss_p}
    \end{subfigure}%
    \begin{subfigure}[b]{0.245\textwidth}
        \centering
        \includegraphics[height=\trainingcurveheight]{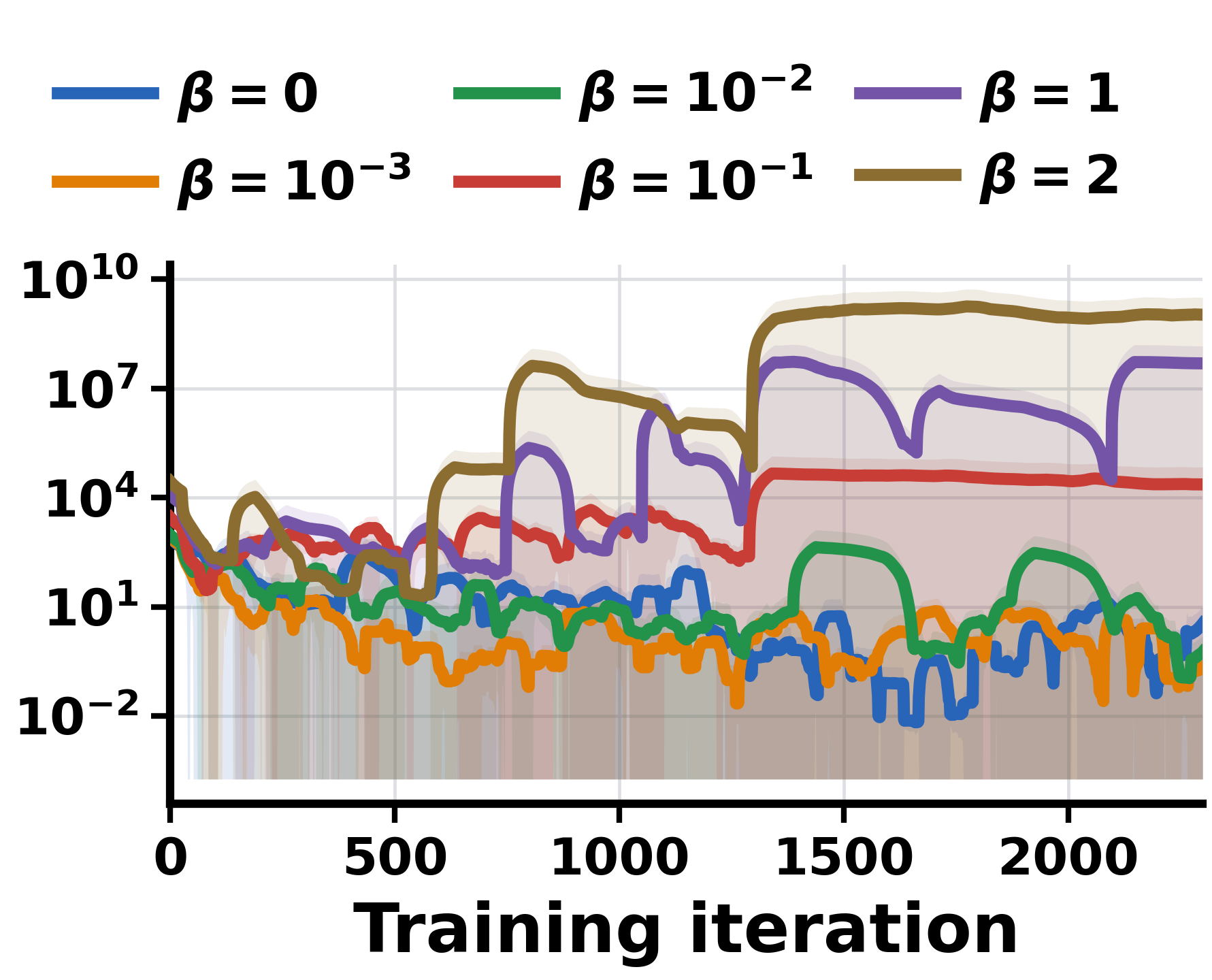}
        \caption{Cartpole}
        \label{fig:training_loss_c}
    \end{subfigure}%
    \begin{subfigure}[b]{0.246\textwidth}
        \centering
        \includegraphics[height=\trainingcurveheight]{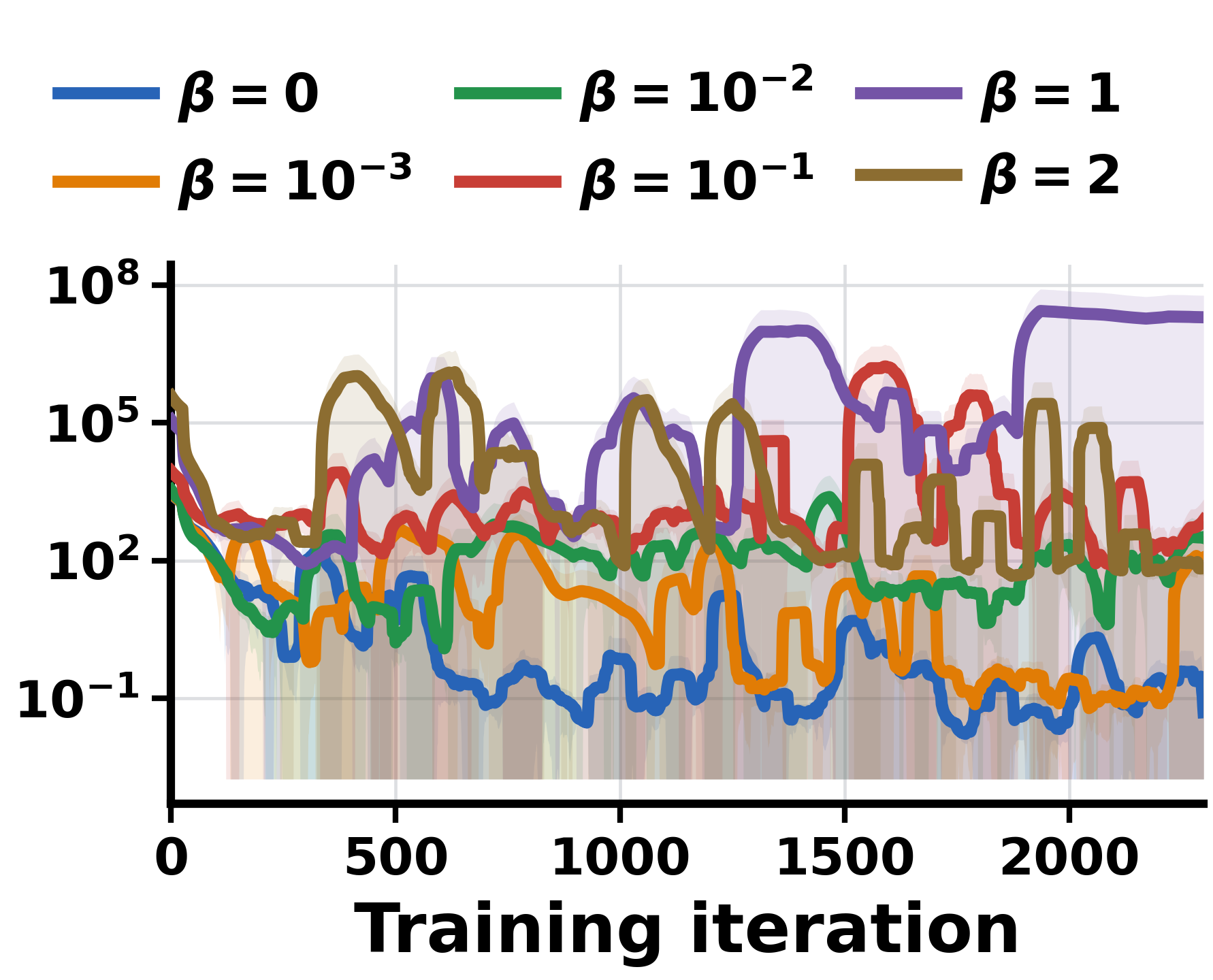}
        \caption{Double Cart Pendulum}
        \label{fig:training_loss_d}
    \end{subfigure}

    \begin{subfigure}[b]{0.262\textwidth}
        \centering
        \includegraphics[height=\trainingcurveheight]{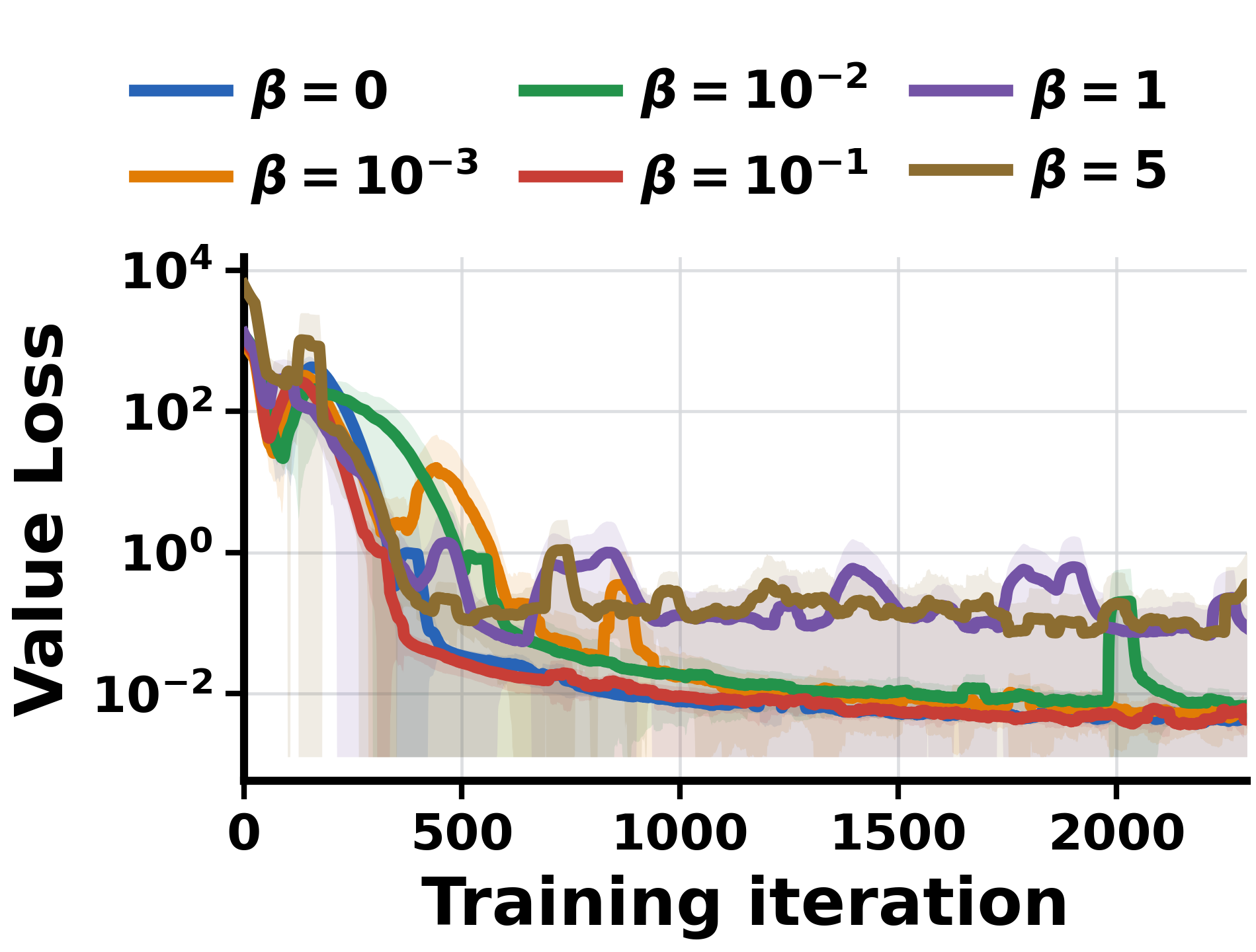}
        \caption{Quadcopter}
        \label{fig:training_loss_q}
    \end{subfigure}%
    \begin{subfigure}[b]{0.247\textwidth}
        \centering
        \includegraphics[height=\trainingcurveheight]{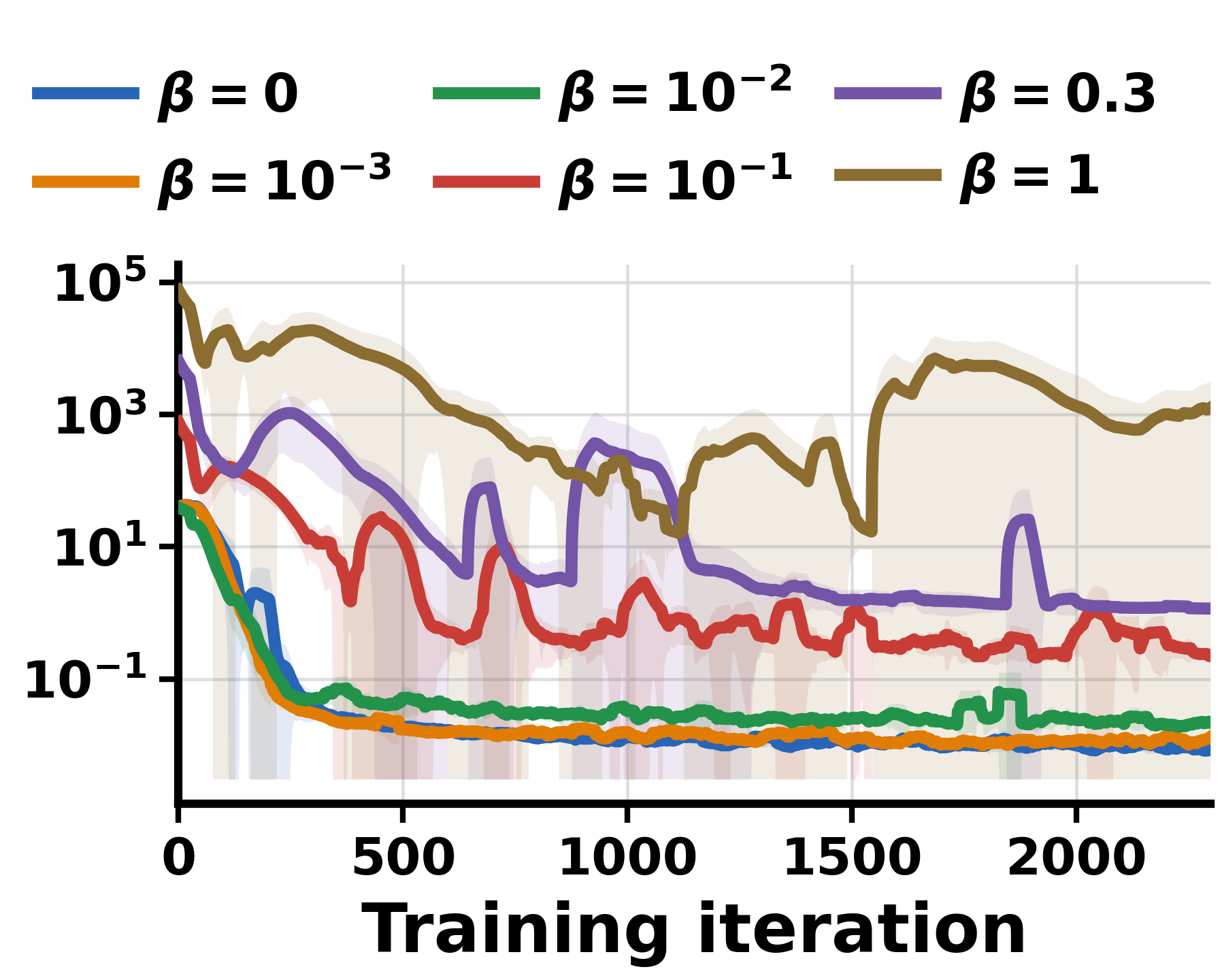}
        \caption{Franka Reach}
        \label{fig:training_loss_f}
    \end{subfigure}%
    \begin{subfigure}[b]{0.241\textwidth}
        \centering
        \includegraphics[height=\trainingcurveheight]{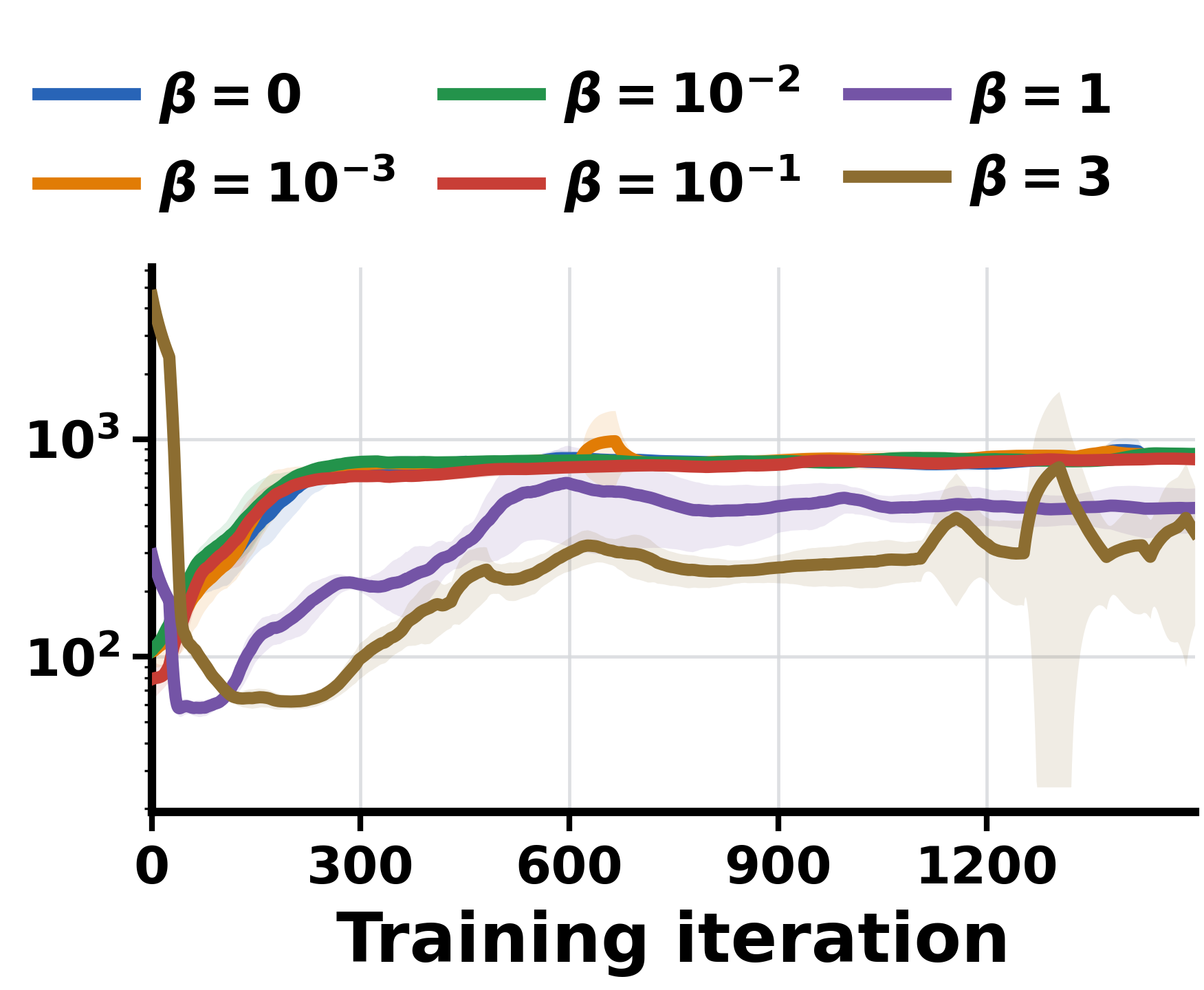}
        \caption{Humanoid}
        \label{fig:training_loss_h}
    \end{subfigure}

    \caption{Training value loss (\Cref{eq:appendix_value_loss}) across the seven Isaac Lab tasks. Each seed is first smoothed with a centered 51-iteration moving average; curves and shaded regions then show the across-seed mean and one standard deviation, respectively.}
    \label{fig:training_curves_loss}
\end{figure*}

The optimization traces in Figures~\ref{fig:training_curves_return} and~\ref{fig:training_curves_loss} agree with the deterministic evaluations in \Cref{fig:beta_sweep}. At $\beta=0$ and at small positive weights, the returns of the pendulum, cart, and Franka tasks rise rapidly and then stabilize.
Increasing $\beta$ generally shifts the return downward, enlarges the across-seed variation, and produces larger or more erratic critic losses, as the unbounded quadratic velocity term changes the scale and variance of the critic target and can dominate the task-progress signal. In contrast, the successful $\beta=0$ curves confirm the trajectory-level argument in \Cref{eq:zeroth_reward_policy_gradient}: future configuration rewards already distinguish actions that brake from actions that overshoot when the actor and critic observe the full state.

Similar training returns across different values of $\beta$ do not imply similar controllers. Each curve optimizes a different scalar objective according to $\beta$, so its numerical value is not a common performance scale across the sweep. A large-$\beta$ policy can obtain a return comparable to that of a successful policy by suppressing velocity while approaching too slowly, stalling outside the target set, or exchanging target progress for a smaller velocity cost.

Moreover, \Cref{eq:appendix_mean_return} averages stochastic training episodes and dense rewards, whereas evaluation requires the deterministic policy to reach, brake, and remain inside a thresholded success set. These different trajectory-component tradeoffs explain why the Quadcopter and Humanoid return curves can occupy a relatively narrow range even when their large-$\beta$ dwell and success rates collapse.

Finally, the increasing Humanoid Value Loss at low $\beta$ is reasonable. \Cref{eq:appendix_value_loss} is an unnormalized absolute squared error against an on-policy, bootstrapped target, not a normalized or held-out prediction error. As the Humanoid learns higher-return trajectories, the magnitude and distribution of $\widehat G_t$ grow and continue to move between PPO iterations; the critic is therefore chasing a nonstationary target on an expanding absolute scale. Its absolute MSE can rise even while its error relative to the return scale falls and policy performance improves. The large-$\beta$ Humanoid runs show the complementary case: their initially large velocity penalties create very large early critic errors, which later decrease as the policy reduces motion, but the resulting conservative behavior still fails the deterministic task.

\section{Numerical Issues When No Velocity Penalties in Trajectory Optimization}
\label{app:hessian}

In direct transcription and  multiple shooting, trajectory optimization discretizes continuous dynamics into a finite-dimensional Non-Linear Program (NLP):
\begin{align}
    \min_{\mathbf{w}} \quad & f(\mathbf{w}) = \sum_{t=0}^{T-1} \left( s_t^\top Q s_t + a_t^\top M a_t \right) + s_T^\top Q_f s_T \\
    \text{s.t.} \quad & g(\mathbf{w}) = 0, \quad h(\mathbf{w}) \le 0,
\end{align}
where $s_t \in \mathbb{R}^{n_s}$ and $a_t \in \mathbb{R}^{n_a}$ denote the state and control vectors at step $t$, respectively. The aggregate decision vector $\mathbf{w} \in \mathbb{R}^{n_w}$ over the horizon $T$ is defined as 
\begin{equation}
    \mathbf{w} = [ s_0^\top, a_0^\top, s_1^\top, a_1^\top, \dots, s_N^\top ]^\top \in \mathbb{R}^{n_w},
\end{equation}
with total dimension $n_w = (T+1)n_s + T n_a$. 

The unconstrained primal Hessian $H_f = \nabla^2_{\mathbf{w}} f(\mathbf{w}) \in \mathbb{R}^{n_w \times n_w}$ possesses a block-diagonal structure:
\begin{equation}
    H_f = \begin{bmatrix}
        Q & 0 & 0 & \dots & 0 \\
        0 & M & 0 & \dots & 0 \\
        0 & 0 & Q & \dots & 0 \\
        \vdots & \vdots & \vdots & \ddots & \vdots \\
        0 & 0 & 0 & \dots & Q_f
    \end{bmatrix}
\end{equation}

If the state weighting matrix $Q \succeq 0$ is positive semi-definite with a nontrivial nullspace $\ker(Q) \neq \{0\}$ (e.g., when velocity penalties are omitted), there exists a nonzero eigenvector $\xi \in \mathbb{R}^{n_s}$ such that $Q \xi = 0$. This induces $T$ linearly independent nullspace directions $\boldsymbol{\xi}_t \in \mathbb{R}^{n_w}$ in the decision space:
\begin{equation}
    \boldsymbol{\xi}_t = \big[ 0_{n_s}^\top, 0_{n_a}^\top, \dots, \underbrace{\xi^\top}_{\text{at } s_t}, \dots, 0_{n_s}^\top \big]^\top
    \quad \implies \quad
    H_f \boldsymbol{\xi}_t = 0.
\end{equation}

Primal-dual interior-point solvers (e.g., IPOPT) compute search directions $(\Delta \mathbf{w}, \Delta \boldsymbol{\lambda})$ by solving the augmented KKT linear system at each iteration:
\begin{equation}
    \begin{bmatrix}
        \nabla^2_{\mathbf{w}\mathbf{w}} \mathcal{L} & J^\top \\
        J & 0
    \end{bmatrix}
    \begin{bmatrix}
        \Delta \mathbf{w} \\
        \Delta \boldsymbol{\lambda}
    \end{bmatrix}
    = -
    \begin{bmatrix}
        \nabla_{\mathbf{w}} \mathcal{L} \\
        g(\mathbf{w})
    \end{bmatrix},
\end{equation}
where $\mathcal{L}$ is the Lagrangian of the NLP, $\boldsymbol{\lambda}$ contains the dual variables, and $J = \nabla_{\mathbf{w}} g(\mathbf{w})$ is the equality constraint Jacobian. When $H_f$ is singular or positive semi-definite, the exact Hessian of the Lagrangian $\nabla^2_{\mathbf{w}\mathbf{w}} \mathcal{L}$ often fails the required inertia condition lacking positive definiteness on the constraint nullspace. The solver must then inject primal diagonal regularization:
\begin{equation}
    \nabla^2_{\mathbf{w}\mathbf{w}} \mathcal{L} \quad \longleftarrow \quad \nabla^2_{\mathbf{w}\mathbf{w}} \mathcal{L} + \gamma_p I_{n_w},
\end{equation}
for some heuristic parameter $\gamma_p > 0$. Excessive regularization distorts the Newton direction, resulting in severe step truncation, oscillatory search behavior, and eventual convergence failure.

\section{Full Derivation of the Trajectory-Level Sufficiency Argument}
\label{app:trajectory_level_view}

This appendix expands the trajectory-level argument of Section~\ref{sec:method}. It explains why RL differs from direct trajectory optimization and why the PPO advantage retains a braking signal even when $r_0$ is velocity-independent, completing the reasoning behind \Cref{eq:closed_loop_kernel}--\Cref{eq:zeroth_reward_policy_gradient}.

\paragraph{RL reshapes trajectory distributions indirectly.}

The policy does not directly optimize or move an individual state trajectory. Instead, each actor update modifies the conditional action distribution $\pi_\Theta$, which perturbs the closed-loop kernel $\mathcal K_\Theta$ of \Cref{eq:closed_loop_kernel} and consequently redistributes probability mass over complete state trajectories $\tau_s$.
The critic and advantage estimator evaluate the relative desirability of sampled trajectory segments, while the actor changes their future likelihood -- redistributing mass over the policy-realizable family $\mathbb P_\Pi$ rather than moving any single trajectory.
Unlike \Cref{eq:direct_trajectory_optimization}, PPO neither manipulates an individual trajectory nor differentiates through the dynamics $p_{\mathrm{dyn}}$. It samples trajectories, estimates returns and advantages, and changes their future probability through the actor. The comparison nevertheless exposes a shared fact: in both cases a reward is accumulated over a dynamically coupled sequence. The direct optimizer makes this coupling explicit in its constraints, while in RL the critic must infer and propagate it from rollouts.

\paragraph{The advantage depends on the action only through $V^\pi$.}

Write $V^\pi(q,v)=\mathbb E_{a\sim\pi(\cdot\mid q,v)}[Q^\pi(q,v,a)]$ for the value function implied by \Cref{eq:zeroth_reward_policy_gradient}. Unrolling its Bellman recursion identifies $V^\pi$ with the expected zeroth-order return-to-go under the closed-loop kernel of \Cref{eq:closed_loop_kernel},
\begin{equation}
    V^\pi(q,v)
    =
    \mathbb E_{\tau_s\sim \mathcal P_\Theta}\left[\textstyle\sum_{k\ge0}\gamma^k r_0(q_{t+k})\;\Big|\;q_t=q,\,v_t=v\right].
    \label{eq:value_as_return_to_go}
\end{equation}
Thus $V^\pi$ genuinely depends on $v$ even though $r_0$ does not, because $v$ enters the dynamics $p_{\mathrm{dyn}}$ and hence shapes the distribution of future configurations $q_{t+1:T}$. Subtracting $V^\pi(q,v)$ from $Q^\pi(q,v,a)$ cancels the shared term $r_0(q)$ exactly, leaving
\begin{equation}
    A^\pi(q,v,a)
    =
    \gamma\Bigl(
        \mathbb E_{s'\sim p_\mathrm{dyn} (\cdot \mid q,v,a)}[V^\pi(s')]
        -
        \mathbb E_{a'\sim\pi(\cdot\mid q,v)}\,\mathbb E_{s'\sim p_\mathrm{dyn} (\cdot \mid q,v,a')}[V^\pi(s')]
    \Bigr).
    \label{eq:advantage_cancellation}
\end{equation}
By \Cref{eq:value_as_return_to_go}, $A^\pi(q,v,a)$ is determined by how $a$ changes the future zeroth-order return relative to the policy average: $a_t\mapsto p_{\mathrm{dyn}}(\cdot\mid q_t,v_t,a_t) \mapsto\mathbb E[V^\pi(s_{t+1})]\mapsto A^\pi_t$. Thus, an action that reduces future configuration error receives positive advantage through this chain, even though the instantaneous reward has no explicit velocity term.

\section{Proof of Theorem~\ref{th:prop1}}
\label{app:proof}

\paragraph{Proof.}
Assume the contrary. By the converse Lyapunov theorem \citep{khalil2002nonlinear}, there exists a continuously differentiable strict Lyapunov function $\mathcal V$ defined on a neighborhood $\mathcal D$ of $x^\ast$, such that $\mathcal V(x^\ast)=0$, $\mathcal V(x)>0$, and $\dot{\mathcal V}(x)=\nabla\mathcal V(x)^\top l_q(x)<0$ for all $x\in\mathcal D\setminus\{x^\ast\}$.

The strict inequality implies that $\nabla\mathcal V(x)\neq 0$ for every $x\in\mathcal D\setminus\{x^\ast\}$, since otherwise $\dot{\mathcal V}(x)=0$. Hence, every $c>0$ is a regular value of $\mathcal V$. Choose $r>0$ such that $\overline{B_r(x^\ast)}\subset\mathcal D$, and let $\delta_r=\min_{x\in\partial B_r(x^\ast)}\mathcal V(x)>0$, where positivity follows from the positive definiteness and continuity of $\mathcal V$. For any $c\in(0,\delta_r)$, define the local sublevel set $\Omega_c=\{x\in\overline{B_r(x^\ast)}:\mathcal V(x)\leq c\}$. The set $\Omega_c$ is compact, and the choice $c<\delta_r$ ensures that $\Omega_c\cap\partial B_r(x^\ast)=\varnothing$; therefore, $\Omega_c\subset B_r(x^\ast)\subset\mathcal D$.

Since $c$ is a regular value, the boundary $\partial\Omega_c=\{x\in B_r(x^\ast):\mathcal V(x)=c\}$ is a compact embedded $C^1$ submanifold of dimension $2n-1$. Moreover, the outward unit normal vector field is well defined by $\hat n(x)=\frac{\nabla\mathcal V(x)}{\|\nabla\mathcal V(x)\|}$. Consequently, for every $x\in\partial\Omega_c$,
\begin{equation}
    l_q(x)^\top \hat n(x)
    =
    \frac{\nabla\mathcal V(x)^\top l_q(x)}
    {\|\nabla\mathcal V(x)\|}
    <0.
\end{equation}
Since $\partial\Omega_c$ is compact and the integrand is continuous and strictly negative on it, its surface integral is strictly negative. On the other hand, using $\nabla\cdot l_q=0$ and applying the divergence theorem gives
\begin{equation}
    0
    =
    \int_{\Omega_c}\nabla\cdot l_q\,dx
    =
    \int_{\partial\Omega_c}l_q^\top\hat n\,dS
    <0,
\end{equation}
which is a contradiction. Therefore, a strict Lyapunov function and local asymptotic stability are impossible and we complete the proof.

\section{An Illustrative Example: Double Integrator}
\label{app:illustr}

\begin{figure}[h!]
    \centering
    \includegraphics[width=1.0\linewidth]{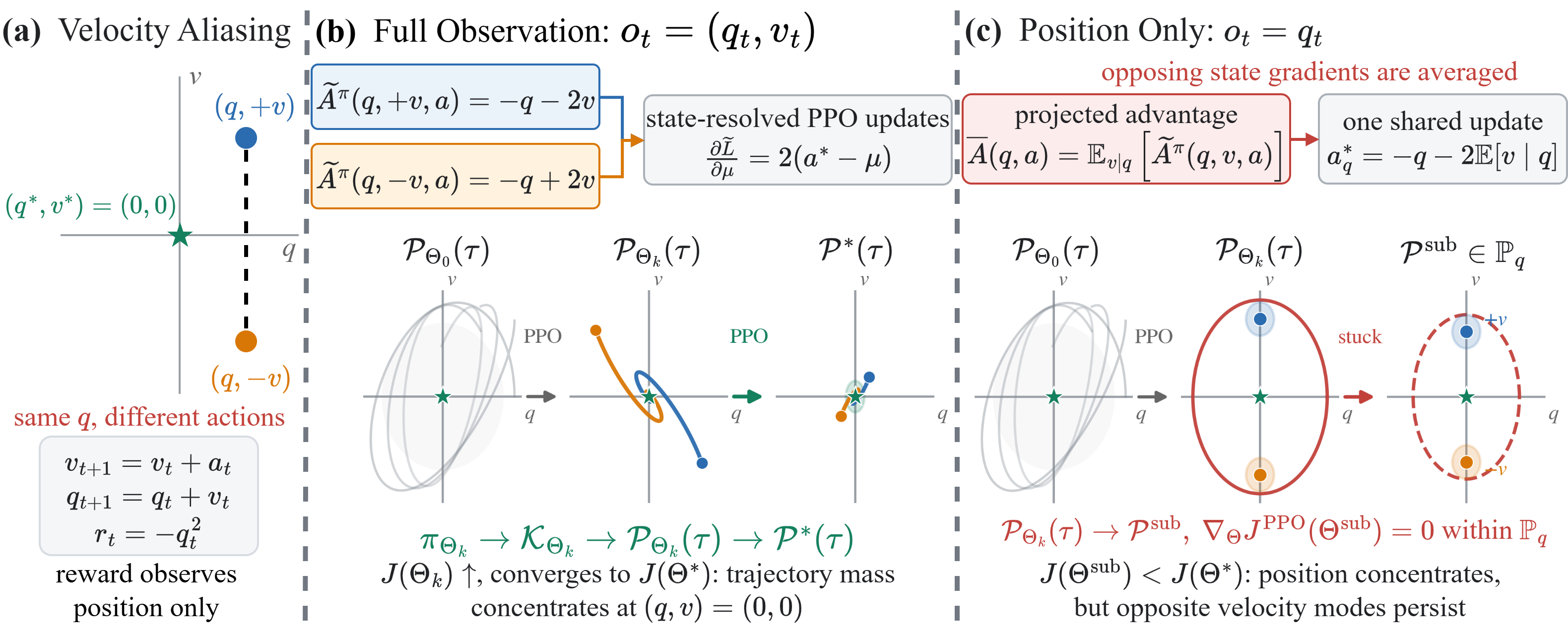}
    \caption{toy example}
    \label{fig:toy_example}
\end{figure}

Consider the discrete-time double integrator
\begin{equation}
    q_{t+1}=q_t+v_t,
    \qquad
    v_{t+1}=v_t+a_t,
    \qquad  
    R_t=-q_t^2,
    \label{eq:toy_dynamics}
\end{equation}
with independent initial states $q_0\sim\mathcal U[-\vartheta,\vartheta]$ and $v_0\sim\mathcal U[-\nu,\nu]$, and objective $J=\mathbb E_{\pi_{\Theta}} [\sum_{t=0}^{T}\gamma^t r_t]$ for $T\geq2$. Because $a_t$ first affects the position at $t+2$, where $q_{t+2}=q_t+2v_t+a_t$, the optimal action is
\begin{equation}
    a_t^\ast=-q_t-2v_t,
    \qquad
    (q_0,v_0)
    \rightarrow(q_0+v_0,-q_0-v_0)
    \rightarrow(0,0)\rightarrow\cdots.
    \label{eq:toy_optimal_policy_trajectory}
\end{equation}
The first two position costs are unavoidable, so $J(\Theta ^ \ast)=-\mathbb E[q_0^2+\gamma(q_0+v_0)^2]$.

To expose the actor update, we retain the first reward term affected by $a_t$. Earlier rewards are action-independent and the omitted factor $\gamma^2>0$ does not change the maximizing action or gradient direction. Thus, the following quantities are the local action-dependent components rather than the complete policy-dependent $Q^\pi,V^\pi$, and $A^\pi$. For a Gaussian policy $\pi(a\mid q,v)=\mathcal N(\mu(q,v),\sigma^2)$,
\begin{align}
    \widetilde Q^\pi(q,v,a)
    &=-(q+2v+a)^2=-(a-a^\ast)^2,
    \\
    \widetilde V^\pi(q,v)
    &=-(\mu-a^\ast)^2-\sigma^2,
    \\
    \widetilde A^\pi(q,v,a)
    &=-(a-a^\ast)^2+(\mu-a^\ast)^2+\sigma^2.
    \label{eq:toy_local_qva}
\end{align}
The PPO mean update is therefore
\begin{equation}
    \frac{\partial\widetilde {\mathcal L}}{\partial\mu}
    =
    \mathbb E_a\left[\widetilde A^\pi\frac{a-\mu}{\sigma^2}\right]
    =2(a^\ast-\mu),
    \label{eq:toy_full_observation_gradient}
\end{equation}
which moves every state-conditioned mean toward $\mu^\ast(q,v)=-q-2v$. The induced trajectory distribution consequently concentrates on \Cref{eq:toy_optimal_policy_trajectory}, although the reward itself contains only $q$.

Now restrict the actor to $\pi_q(a\mid q)=\mathcal N(\mu_q(q),\sigma^2)$. Maximizing the local objective averaged over hidden velocities gives $\mu_q^\ast(q)=-q-2\mathbb E[v\mid q]$. At this shared mean, the state-resolved update and its observation-conditioned average are
\begin{equation}
    \frac{\partial\widetilde {\mathcal L}}{\partial\mu}(q, v)
    =2[a^\ast(q,v)-\mu_q(q)],
    \qquad
    \mathbb E\left[\frac{\partial\widetilde {\mathcal L}}{\partial\mu} (q,v)\mid q\right]=0,
    \label{eq:toy_gradient_cancellation}
\end{equation}
which moves the mean toward $\mu^\ast(q,v)=-q-2\mathbb E[v\mid q]$. Thus, PPO becomes ``stuck" at a suboptimal trajectory distribution $\mathcal P ^ {\mathrm{sub}} \in \mathbb P_q $: it is stationary after averaging over states that share an observation, even though their state-resolved gradients are nonzero and require different braking actions. The irreducible deterministic action error is $\mathbb E[(\mu_q^\ast-a^\ast)^2\mid q]=4\operatorname{Var}(v\mid q)>0$ whenever the hidden velocity is ambiguous. The zeroth-order reward still identifies the better trajectories, but the zeroth-order actor cannot realize their required state-dependent redistribution.

\end{document}